\documentclass[letterpaper]{article} % DO NOT CHANGE THIS
\usepackage[preprint]{aaai2027}  % arXiv version: real authors shown, no copyright footer
\usepackage[hyphens]{url}  % DO NOT CHANGE THIS
\usepackage{graphicx} % DO NOT CHANGE THIS
\usepackage{natbib}  % DO NOT CHANGE THIS AND DO NOT ADD ANY OPTIONS TO IT
\usepackage{caption} % DO NOT CHANGE THIS AND DO NOT ADD ANY OPTIONS TO IT
\usepackage{booktabs}

\usepackage{amsmath}
\newtheorem{proposition}{Proposition}

\title{The Null Token Knows: Reducing Message-Free Hallucination in ASR and NMT}

\author{
    Kirill Borodin\corresponding\textsuperscript{\rm 1,\rm 2,\rm 3},
    Vasiliy Kudryavtsev\textsuperscript{\rm 1,\rm 2,\rm 3},
    Ivan Viakhirev\textsuperscript{\rm 4,\rm 5},
    Grach Mkrtchian\textsuperscript{\rm 1,\rm 3}
}
\affiliations{
    \textsuperscript{\rm 1}Lab260, Moscow, Russia \quad
    \textsuperscript{\rm 2}BitmanagerAI, Dubai, UAE \quad
    \textsuperscript{\rm 3}MTUCI, Moscow, Russia \\
    \textsuperscript{\rm 4}ITMO, Saint Petersburg, Russia \quad
    \textsuperscript{\rm 5}FRC RAS, Saint Petersburg, Russia \\
    kborodin.research@gmail.com
}

\begin{document}

\maketitle

\begin{abstract}
Modern encoder--decoder systems can produce fluent text even when their
input contains no recoverable message. We study this failure in ASR and NMT
through the models' reserved null tokens, asking whether the score for
ending generation already carries a usable abstention signal. Across speech
recognizers and translation models, we audit native null-token scores and
scalar logit shifts. In Whisper, we additionally probe decoder states and
compare supervised row edits with conventional external gates. The evaluated
models often expose a useful abstention signal, but stock decoding does not
reliably act on it. Raising the null-token score can sharply suppress
fabrication, but aggressive intervention also deletes valid speech or
shortens legitimate translations. These findings turn the null token into
a diagnostic lens on hallucination and motivate evaluating abstention
methods by both suppression and deletion costs, rather than by hallucination
reduction alone.
\end{abstract}

% Uncomment to link to code / datasets / extended version.
% \begin{links}
%     \link{Code}{https://...}
% \end{links}

\section{Introduction}

Attention-based encoder--decoder (AED) systems can fabricate fluent output
from inputs that carry no message. Whisper \citep{radford2023whisper}
transcribes confident sentences from silence, room tone, and music, with
documented downstream harms \citep{koenecke2024careless,
frieske2024hallucinations}; neural translators can likewise return fluent
text for degenerate, message-free sources \citep{lee2018hallucinations,
guerreiro2023needle, raunak2021curious}. This failure does not reliably
disappear with scale: across six Whisper checkpoints, our developmental
endpoint is non-monotonic even as WER falls
(Table~\ref{tab:zoo}) \citep{viakhirev2026spectral}.

Operational systems counter this failure with native no-speech scores,
VAD/endpointing, and reject options
\citep{radford2023whisper,fireredvad2025,chow1970optimum}. Other work
re-trains attention heads \citep{calmwhisper2025}, steers encoder features
through sparse autoencoders \citep{audiosae2026}, or distills against a
clean teacher \citep{llt2025}. Yet each evaluated family already exposes a
reserved output coordinate: EOT in AED decoders, EOS in translation, and
the semantically different alignment blank in CTC/RNN-T
\citep{graves2006ctc,graves2012rnnt}. Because Whisper encoders also
represent non-speech events \citep{gong2023whisperat}, we ask a narrower
question: what battery-specific separation appears in decoder states and
native scores, and what fabrication--deletion tradeoff follows from editing
the reserved coordinate?

\begin{figure}[t]
\centering
\includegraphics[width=0.90\columnwidth]{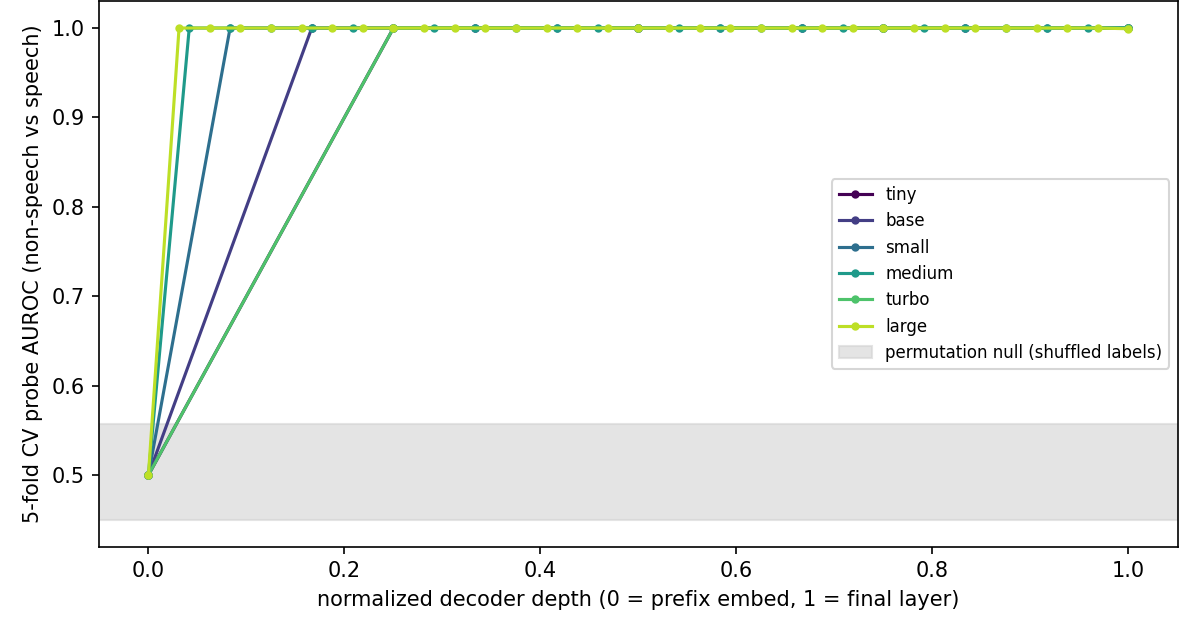}
\caption{Developmental clip-stratified probes separate non-speech from
speech after the first Whisper decoder block. The layer-0 embedding is at
chance; the gray band is the permutation baseline. Folds are not grouped
by source corpus, so this does not establish source-independent encoding.}
\label{fig:layersep}
\end{figure}

On the developmental battery, frozen decoder states linearly separate
non-speech from speech after the first decoder block
(Figure~\ref{fig:layersep}), and native null-token scores distinguish the
two conditions even when the null token loses the output decision. We
therefore use readout edits only as instruments for moving the abstention
operating point. A scalar bias shifts constant prior odds; the analytic
graft and trained row are supervised, state-dependent variants of one
output coordinate. Their results do not establish an advantage over a
scalar or external gate. Prior shift is a conditional model for the scalar
case \citep{saerens2002, lipton2018bbse}; we test where its operating-point
predictions hold.

Our evidence comes from a developmental 1060-clip stress battery, a
source-disjoint locked Whisper-small 300/300 evaluation, and a 15-model zoo
spanning AED recognition, CTC/RNN-T, and neural translation. Three
questions guide the analysis.
\begin{itemize}
\item \textbf{RQ1 (battery separation and readout).} How do decoder states
and native null-token scores separate the developmental conditions across
layers and positions (Section~\ref{sec:readout})?
\item \textbf{RQ2 (checkpoint audit).} How do scalar and row edits trade
empty-target output against speech deletion, how do frozen checkpoints rank
on the locked sensitivity grid, and where does the conditional prior-shift
account fail (Section~\ref{sec:calibration})?
\item \textbf{RQ3 (boundary tests).} Which findings fail to transfer across
architectures, modalities, languages, and long-form audio, and when are
standard gates or segmentation safer (Sections~\ref{sec:class}
and~\ref{sec:limits})?
\end{itemize}

\paragraph{Measurement boundary.}
We measure deterministic \emph{normalized lexical output} on spans with
empty lexical targets. After removing documented non-lexical event tags,
punctuation, and digits, any retained letter is fabricated by construction.
The empty-target label is exact, but the score is normalizer-dependent and
is not semantic adjudication. We exclude fabrication interleaved with
genuine content, where the target is non-empty and no clean detector
exists. WER cannot distinguish fabricated from misrecognized words, just
as reference-overlap metrics miss generation faithfulness
\citep{maynez2020faithfulness, ji2023survey}. Although documented harms
include interleaved fabrication \citep{koenecke2024careless}, our claims
cover only whole-span empty lexical targets.

\section{Background and Related Work}

ASR and NMT hallucination is well documented
\citep{koenecke2024careless,baranski2025investigation,
lee2018hallucinations,guerreiro2023needle}. Native no-speech scoring and
VAD/endpointing are the closest operational baselines
\citep{radford2023whisper,fireredvad2025}; selective prediction formalizes
the same reject option \citep{chow1970optimum,elyaniv2010foundations}.
Complementary Whisper mitigations edit attention heads, encoder features,
teachers, or emitted text
\citep{calmwhisper2025,audiosae2026,aparin2026steering,llt2025}.
Whisper representations are already known to encode non-speech events
\citep{gong2023whisperat,viakhirev2026spectral}; our distinct question is whether the decoder's
reserved null-token pathway carries a useful signal.

Under label shift, the Bayes decision changes by a closed-form prior-odds
term \citep{saerens2002,lipton2018bbse}. NMT models are miscalibrated, with
EOS especially affected \citep{kumar2019calibration}; inference
miscalibration also persists under generated context
\citep{wang2020inference}. We combine those results with frozen-state probes
and logit-lens readouts
\citep{alain2017probes,belinkov2022probing,belrose2023tunedlens} to
characterize the available abstention signal. The intervention uses an
in-vocabulary reject option
\citep{chow1970optimum,elyaniv2010foundations}: a one-coordinate
vector-scaling calibrator, whereas temperature scaling cannot change a
greedy argmax \citep{guo2017calibration}.

Translation alternatives alter search with coverage, length, or lexical
constraints \citep{wu2016gnmt,murray2018correcting,post2018fast}. Our edit
changes the null-token scoring function. These alternatives are related
inference-time methods, not matched baselines in our experiments. We test
greedy ASR and each NMT backend's fixed decoder; beam-width invariance and
continuous long-form inference are outside the claim.

The established ingredients are rejection, label-shift correction, EOS
calibration, native no-speech scoring, and VAD. Our empirical contribution
is a diagnostic and falsification audit: it measures what direct
null-coordinate edits can control, records their post-freeze
fabrication--deletion cost among five Whisper-small checkpoint conditions,
and identifies where stronger representation, prior-shift, risk-frontier,
and cross-family hypotheses fail. Native and external rejectors are
compared only on inspected developmental data. We do not claim to discover
abstention or one shared mechanism for EOT, EOS, and CTC/RNN-T blank.

\section{Experimental Setup}

\paragraph{Data and metric.}
Our stress battery contains $630$ non-speech and $430$ speech clips
($3.1$ h total): synthetic silence/room tone; MUSAN noise/music
\citep{snyder2015musan}; ESC-50 \citep{piczak2015esc}; UrbanSound8K
\citep{salamon2014urbansound}; and clean/degraded FLEURS speech
\citep{conneau2023fleurs}. A deterministic filename-based assignment
partitions MUSAN into disjoint calibration and evaluation halves. FLEURS
test and LibriSpeech test-clean \citep{panayotov2015librispeech} are used
only for evaluation (supplementary Table~\ref{tab:stress}). For
empty-reference clips, the endpoint is non-empty lexical output after the
documented normalizer removes event tags, punctuation, and digits. It is
deterministic given that normalizer, but does not count tag-only,
number-only, or wrong-script output hidden by normalization. We call its
clip incidence the \emph{normalized lexical-output rate} (LOR), reserving
``hallucination'' for the broader failure class.
Headline rates have clip-level Wilson intervals \citep{wilson1927probable}; paired
\texttt{eot\_row} effects use exact McNemar tests
\citep{mcnemar1947note} with Holm correction
(supplementary Tables~\ref{tab:stats}--\ref{tab:mcnemar}).
These finite-battery summaries are not cluster-robust or estimates of
source-level transfer.

\paragraph{Models and interventions.}
The core zoo has six Whisper scales \citep{radford2023whisper}, Canary-1B
\citep{puvvada2024canary}, OWSM v3.1 \citep{peng2024owsm31}, five CTC and
two RNN-T recognizers, plus NLLB-200 and MarianMT
(supplementary Table~\ref{tab:zoo}). Additional stock and readout
checks cover Distil-Whisper, SeamlessM4T-v2-large, and
Parakeet-TDT-1.1B. Their model-specific endpoints are reported in the
supplement; no intervention claim is made for these checks.

Every stock/patch pair uses the same decoder. We distinguish three
interventions. A scalar null-token bias adds a constant to one logit. An
analytic row graft fits a supervised linear direction from labeled hidden
states and adds it to one unembedding row. A trained EOT row updates the
parameters of one output coordinate on a balanced calibration mix.

The supplement gives the common optimization recipe, regime matrix, run
counts, data partitions, and pipeline checks. It also inventories the code,
manifests, configurations, learned rows and biases, selection logs, and
per-example outputs prepared for release.

\paragraph{Analysis and selection pools.}
The stress battery and reduced subsets were inspected during method
development, so their intervals quantify within-set clip variation rather
than method adaptivity. After freezing methods and configurations, we
evaluated once on 300 non-speech and 300 speech clips whose audio hashes,
source recordings, and speakers do not overlap fitting or selection pools.
The non-speech side draws from an unused silence seed and unused
MUSAN, ESC-50, UrbanSound8K, and music recordings; speech covers clean,
low-SNR, WHAM!-degraded, and accented inputs. A collision checker rejects a
deliberately contaminated manifest. Test-clean had appeared previously as
a reported benchmark, so the speech side is source-disjoint but not
``virgin''; the unavailable spontaneous stratum is recorded as absent.
Source disjointness addresses overlap, not deployment distribution shift.

We compare methods in two ways. On the developmental common-cost set, an
operating point is reported as within an illustrative cap when at most
$10\%$ of real-speech clips are emptied; this threshold is not an
application-derived utility. On the locked set, the pre-specified
sensitivity family,
frozen before decoding, is
\[
\begin{aligned}
C_\kappa
&=\text{fabricated lexical words/min}\\
&\quad+\kappa\,\text{deleted reference words/min},\\
\kappa&\in\{0.5,1,2,5,10\}.
\end{aligned}
\]
The weight $\kappa$ combines relative exposure/prevalence and per-word
harm; $C_\kappa$ is a conditional-rate sensitivity index, not total WER or
an expected deployment cost. The locked Whisper-small evaluation is the only
post-freeze reporting pool; all zoo, probe, NMT, multilingual, and
long-form results are developmental or boundary-setting.
The full data-role ledger and source audit are in supplementary
Table~\ref{tab:decision-record} and
Section~\ref{sec:extended-validation}.

\section{RQ1: Developmental Separation and Readout}
\label{sec:readout}

RQ1 measures how decoder states and native output scores separate the
assembled developmental conditions. It also examines whether a decoding
rule or later sequence positions account for the observed output.

\subsection{Linear Recoverability in Decoder States}

For each decoder layer $\ell$ we take the hidden state at the first content
position (after the forced prefix) on the 1060-clip stress battery, and fit
a logistic probe with 5-fold cross-validation to separate non-speech from
speech. Layer 0 is the prefix-token embedding before any decoder block, and
its probe is at chance. From the first decoder block onward, the
clip-stratified probe reaches the maximum observed AUROC on all six scales
(Figure~\ref{fig:layersep}; Table~\ref{tab:readout}, first column).

A permutation null (shuffled labels) sits at $0.50$ and a top-50 PCA
probe reproduces the result, reducing concern about high-dimensional
overfit. These probes establish linear recoverability on the assembled
battery. Because folds are not grouped by source corpus, they do not by
themselves establish a source-independent representation or a causal
failure location.

\begin{figure*}[t]
\centering
\includegraphics[width=0.98\textwidth]{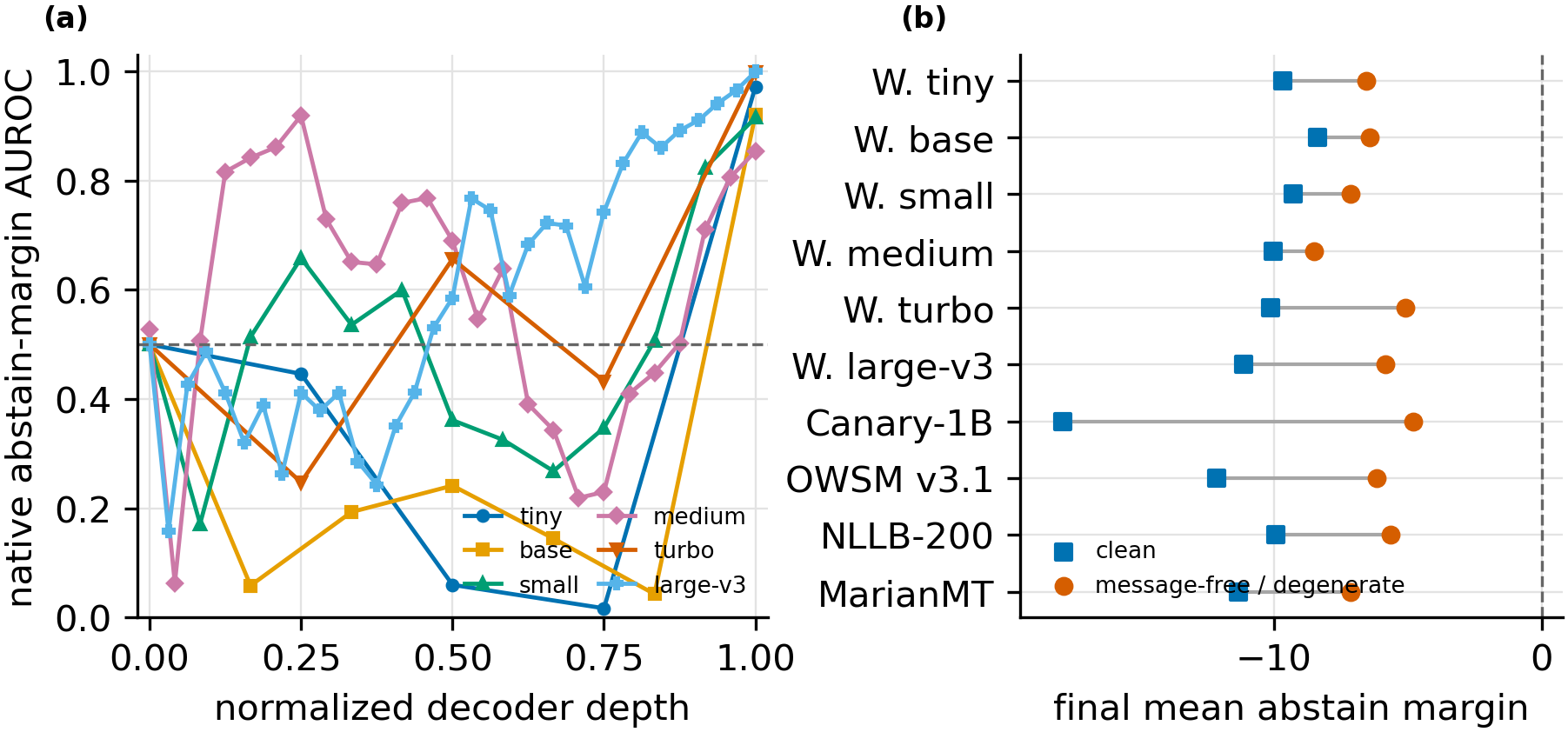}
\caption{Developmental native null-token readout at decode step~0.
(a) Whisper abstain-margin AUROC across the 1060-clip battery and decoder
depth. (b) Class-conditional mean final margins on each model's named
developmental subsets; connecting lines join the two means and zero is the
abstention boundary. Subset denominators are in the supplement.}
\label{fig:logitlens-summary}
\end{figure*}

The supervised probe measures in-battery linear recoverability. A complementary
logit lens \citep{nostalgebraist2020logitlens, belrose2023tunedlens} applies
the model's frozen final LayerNorm and unembedding at each depth. Unlike the
trained probe, this native abstain direction is non-monotone through the
stack (Figure~\ref{fig:logitlens-summary}a). At the final readout, the
message-free margin exceeds the clean-input margin in every evaluated
model but remains below the abstention boundary
(Figure~\ref{fig:logitlens-summary}b). The native coordinate therefore
orders these evaluated conditions without selecting abstention. Full
cross-architecture traces, including Marian's distinct late-layer
trajectory, appear in supplementary
Section~\ref{sec:logitlens-detail}.

% Source: SLENSER0/asr_hallucinations, commit 7976db566af26d7aa594eb23dd1b419f9599ba95.

\subsection{Native Null-Token Readout}

Linear probe performance does not determine the model's argmax, so we also
measure its native output scores. The
$\langle\mathrm{nospeech}\rangle$ and EOT logits distinguish non-speech
from speech across Whisper scales (Table~\ref{tab:readout}). The same
step-0 ordering appears on named developmental subsets for the evaluated
non-Whisper AED and translation models: the abstain margin
$\mathrm{logit}(\mathrm{EOT})-\max_t\mathrm{logit}(\mathrm{content}_t)$
distinguishes the evaluated conditions (supplementary
Table~\ref{tab:scalar-summary}). Distil-Whisper
provides an additional readout check in supplementary
Table~\ref{tab:zoo}.

On message-free inputs, the null-token score often remains below a content
token. This gap motivates testing the output coordinate as a calibration
target (Section~\ref{sec:calibration}); it does not establish why the stock
checkpoint learned that score.

\begin{table}[t]
\centering
\footnotesize
\setlength{\tabcolsep}{1.5pt}
\begin{tabular}{lcccccc}
\toprule
 & state & \multicolumn{2}{c}{AUROC@0} & \multicolumn{2}{c}{row/ns geometry} & graft \\
\cmidrule(lr){2-2}\cmidrule(lr){3-4}\cmidrule(lr){5-6}
Model & probe & ns & EOT & $\cos(\Delta,w_{\rm ns})$ & random $p_{95}$ & ns$\to$EOT \\
\midrule
tiny     & 1.00 & 0.953 & 0.977 & $+0.119$ & 0.103 & fails \\
base     & 1.00 & 0.963 & 0.840 & $+0.043$ & 0.092 & fails \\
small    & 1.00 & 0.878 & 0.835 & $+0.114$ & 0.112 & fails \\
medium   & 1.00 & 0.915 & 0.728 & $+0.034$ & 0.026 & fails \\
turbo    & 1.00 & 0.988 & 0.995 & $-0.014$ & 0.046 & fails \\
large-v3 & 1.00 & 0.814 & 0.999 & $-0.032$ & 0.046 & fails \\
\bottomrule
\end{tabular}
\caption{Developmental Whisper state-probe and native-logit AUROC at
step~0. ``ns'' is Whisper's no-speech token and EOT is end-of-transcript.
$\Delta$ is the trained EOT-row update; $\cos(\Delta,w_{\rm ns})$ compares
it with the native no-speech row, and random $p_{95}$ is the 95th
percentile of matched random-direction cosines. The final column tests
grafting the no-speech row direction into EOT. Cross-architecture readouts are in
supplementary Table~\ref{tab:scalar-summary}.}
\label{tab:readout}
\end{table}

\subsection{Native and External Abstention Scores}

The trained update $\Delta$ is weakly aligned with the
$\langle\mathrm{nospeech}\rangle$ direction. Its cosine is comparable to
the upper tail of random directions on every model
(Table~\ref{tab:readout}). Grafting the no-speech direction into the EOT row,
$w_{\mathrm{EOT}}\!\leftarrow\!w_{\mathrm{EOT}}+\lambda(w_{\mathrm{ns}}-w_{\mathrm{EOT}})$,
does not suppress normalized lexical output.

Whisper defines \texttt{no\_speech\_prob} at the SOT position, where the
no-speech token competes with language tokens. Read there by one shared
implementation, it leaves $4.3\%$ residual LOR on the developmental
battery under the illustrative $10\%$ speech-side cap, versus $89.8\%$
for thresholding the native EOT margin. Thus the separate native detector
is useful; the observed failure is specific to the EOT operating point,
not an absence of all native abstention signal.

On an identical Whisper-small common-cost set, a frozen-state logistic gate
reaches $0.0\%$ LOR and FireRedVAD \citep{fireredvad2025} reaches
$1.3\%$, both within the cap. A constant EOT bias is also within the cap
($24.3\%$ LOR; $7.4\%$ real-speech empty output), whereas the
trained row ($0.2\%$; $67.7\%$) and analytic graft ($0.0\%$; $35.6\%$)
operate beyond it. The row edits package a state-dependent score
change in one checkpoint row, but that is an implementation format rather
than a risk-frontier advantage. All arms and denominators appear in
supplementary Table~\ref{tab:common-cost}.

\begin{table}[t]
\centering
\scriptsize
\setlength{\tabcolsep}{3pt}
\begin{tabular}{lrr}
\toprule
Developmental method & LOR & speech rejected/empty \\
\midrule
native EOT margin & 89.8 & $\le10.0$ \\
native \texttt{no\_speech\_prob} & 4.3 & $\le10.0$ \\
frozen-state logistic gate & 0.0 & $\le10.0$ \\
FireRedVAD front-end & 1.3 & $\le10.0$ \\
\midrule
stock checkpoint & 96.2 & 0.0 \\
EOT bias $b=5$ & 24.3 & 7.4 \\
trained EOT row & 0.2 & 67.7 \\
analytic graft $\lambda=2$ & 0.0 & 35.6 \\
\bottomrule
\end{tabular}
\caption{Matched but inspected Whisper-small comparison on 630 non-speech
and 430 speech clips (percent). For the four thresholded detectors, the
review bundle preserves only that the selected point satisfies the
$\le10\%$ cap, not the exact speech-side rate. The cap is illustrative and
rows are descriptive points, not efficacy or superiority estimates.}
\label{tab:common-cost-main}
\end{table}

\subsection{Parameter-Locus Controls}

Tuning LayerNorm can suppress message-free output, but a matched-capacity
random subnetwork can do so as well across the evaluated learning-rate and
seed grid (supplementary Table~\ref{tab:lrgrid}). Their speech costs
differ and are sensitive to optimization, especially for Whisper medium.
The control therefore supports reachability from several parameter subsets,
not a privileged locus or equivalent safety.

\subsection{Sequence-Level Behavior}
\label{sec:dynamics}

\paragraph{Configuration suppression.}

Removing the stock rule that suppresses EOT at the first output position
does not change stock LOR. The rule can block a calibrated
EOT decision, however, so all stock and patched comparisons use the same
configuration. Details and the diagnostic control are in
supplementary Section~\ref{sec:pipeline}.

\paragraph{Position-wise margins.}

To see where along the sequence the failure lives, we teacher-force each stock
model on its own saved hallucinated transcript over non-speech audio and read
the abstain margin per position
(supplementary Table~\ref{tab:ratchet}). Stock margins are negative at
most measured positions. The calibrated readout increases the margin most
at early positions, but turbo and large-v3 remain negative later in the
sequence. This result limits the intervention to short, whole-segment
abstention. The separate long-form test reduces gap words but increases
WER. Correctly cutting speech regions with VAD, rather than merely muting
gaps, dominates the row edit on the evaluated long-form set
(supplementary Table~\ref{tab:longform-corrected}).

\section{RQ2: Null-Coordinate Diagnostic Audit}
\label{sec:calibration}

We treat prior shift as a candidate explanation for the step-0 null-token
margin, derive its conditional scalar prediction, and test where that
explanation fails. We then compare the scalar intervention with two
supervised row-based recalibrators.

\subsection{Conditional Prior-Shift Model}

Under label shift, a scalar correction is exact only if the neural margin
is calibrated unit-scale posterior log odds. With fixed
class-conditionals, that assumption gives
$m_{\rm test}=m_{\rm train}+\mathrm{logit}(\pi_{\rm test})
-\mathrm{logit}(\pi_{\rm train})$ \citep{saerens2002,lipton2018bbse}.
An arbitrary EOT-versus-content margin need not satisfy it; otherwise the
scalar is merely a one-coordinate recalibrator. The full proposition,
assumptions, and derivation are in supplementary
Section~\ref{sec:prior-shift}.

Unit base-rate slope and a margin-predicted error-balance bias are therefore
applicability checks, not established properties of stock decoders. They
neither identify the unknown training prior nor show that prior shift caused
stock behavior.

\subsection{Operating-Point Prediction}

A bias $\beta$ on the null-token logit makes the model abstain at step 0
exactly where the abstain margin exceeds $-\beta$, so the \emph{whole}
dose-response curve is predictable from the logged step-0 margin
distribution before another decode. We define $\beta^\star$ as the
evaluated grid point that minimizes the absolute difference between missed
abstentions and false abstentions. It corresponds to the shifted Bayes
threshold only under the additional symmetric-margin and equal-cost
assumptions.

The margin prediction agrees with Canary at grid resolution and is adjacent
to the OWSM empirical point, but it fails on both translation models
(supplementary Table~\ref{tab:scalar-summary} and
Figure~\ref{fig:betastar}). Thus the scalar
null-coordinate intervention is effective on these evaluated systems, but
the no-fit operating-point predictor is not general. The same fixed subset
is used to choose and report each bias, so these dose-response results are
descriptive rather than locked-test estimates.

\subsection{Base-Rate Dependence of Learned Margins}

We train each Whisper EOT row at seven
non-speech fractions $p\in[0.1,0.9]$ and read the converged margin
(supplementary Table~\ref{tab:baserate-replicated}). Every per-scale slope is
positive and all unadjusted per-scale intervals exclude zero. Some scales include
unit slope while others are steeper.
A random-effects combination gives slope $1.29$ with $95\%$ CI
$[0.94,1.65]$, with substantial heterogeneity ($I^2=84\%$). Equivalence to
unit slope under the pre-specified $[0.8,1.2]$ band does not hold, and the
random-effects scale trend is not established ($p=0.118$). Seed
replication comprises 18 independently resampled runs and 126 row
trainings. These developmental training runs are consistent with positive
log-odds dependence, but do not establish a universal unit coefficient,
the stock checkpoints' training priors, or out-of-source generalization.
The earlier five-point small sweep is retained in
supplementary Table~\ref{tab:baseratepoints} for the row-norm
diagnostic.

\subsection{Null-Token Interventions}

The trained EOT row is a supervised update to one vocabulary coordinate,
not one scalar. It changes hundreds of row parameters. At the common
step-50 checkpoint it suppresses message-free output across the full
battery (supplementary Table~\ref{tab:fix-summary}); confidence
intervals and paired tests are in
supplementary Tables~\ref{tab:stats}--\ref{tab:mcnemar}.
Checkpoint trajectories and first-zero comparisons are adaptive development
analyses, not independent test estimates.

The analytic row graft is also supervised and state-dependent. A logistic
probe supplies a direction $w$, which is added to the EOT row with scale
$\lambda$. Non-speech calibration clips impose lower bounds on $\lambda$,
while speech clips impose upper bounds. Their intersection defines an
admissible interval, and the reported scale is an interior point of that
interval. Supplementary Section~\ref{sec:lambda} gives pseudocode for
the complete procedure.

The analytic calibration pools are disjoint from the reported stress
battery and LibriSpeech test-clean. Its clean-speech constraints do not,
however, certify a new acoustic distribution. The method performs well on
five Whisper scales and regresses on medium (supplementary
Table~\ref{tab:fix-summary}). The
analytic and trained row directions are weakly aligned, so they should be
viewed as different recalibrators of the same output coordinate.

Additional controls compare random output coordinates, EOT suppression,
loss weighting, and a full-vocabulary affine calibrator. They support the
choice of the null-token coordinate on the evaluated battery; full results
are provided in the supplement.

On the inspected battery, the common step-50 row reaches zero LOR on all six
Whisper scales; held-out probe WER is unchanged on four scales and rises on
turbo and large-v3. The disjointly fitted analytic graft leaves
$0.3$--$8.4\%$ LOR and has sharply model-dependent test-clean cost, including
$2.88\to7.60$ WER on medium. The complete developmental table, including
admissible intervals and denominators, is supplementary
Table~\ref{tab:fix-summary}.

\subsection{Locked Fabrication--Deletion Sensitivity}

Table~\ref{tab:locked-cost} evaluates the frozen Whisper-small methods on
the post-freeze, source-disjoint 300/300 set. Among these five checkpoint
conditions, scalar bias $b=5$ has the lowest cost point estimate on this
sample at $\kappa=0.5,1$; stock has the lowest point estimate for
$\kappa\ge2$.
The trained row never minimizes the displayed grid. It nevertheless moves
the measured endpoint sharply. LOR falls
from $91.7\%$ to $3.3\%$, but the intervention incurs
$33.05$ deleted reference words/min versus stock's $1.58$. Its false
silence is $8.0\%$ on clean speech, $42.7\%$ at low SNR, and $38.7\%$ on
accented speech. These group rates characterize only the trained row and do
not support a comparative group-safety claim. The observed ranking reversal
shows sensitivity to the displayed weights; sampling uncertainty for cost
gaps and crossover location was not quantified. External gates were not
decoded on this locked set, so this is neither a superiority test nor a
ranking of the best available rejectors.

\begin{table*}[t]
\centering
\scriptsize
\setlength{\tabcolsep}{3.2pt}
\begin{tabular}{lrrrrrrr}
\toprule
Method & $F$ & $D$ & $C_{0.5}$ & $C_1$ & $C_2$ & $C_5$ & $C_{10}$ \\
\midrule
stock & 32.5 & 1.6 & 33.3 & 34.1 & \textbf{35.7} & \textbf{40.4} & \textbf{48.3} \\
trained EOT row & 4.3 & 33.0 & 20.8 & 37.3 & 70.4 & 169.5 & 334.8 \\
EOT bias $b=5$ & 10.1 & 17.6 & \textbf{18.9} & \textbf{27.7} & 45.3 & 98.0 & 185.7 \\
EOT bias $b=6$ & 6.4 & 34.6 & 23.7 & 41.0 & 75.8 & 179.9 & 353.6 \\
analytic graft $\lambda=2$ & 35.4 & 17.4 & 44.1 & 52.8 & 70.1 & 122.1 & 208.8 \\
\bottomrule
\end{tabular}
\caption{Locked conditional-rate sensitivity
$C_\kappa=F+\kappa D$ (lower is better). $F$ is total words retained by
the specified lexical normalizer divided by total non-speech audio minutes;
$D$ is total reference-word deletions divided by total speech audio
minutes. Component values are
rounded to one decimal; displayed costs come from unrounded values. Methods
and weights were author-frozen before this 300/300 decode. Boldface marks
the descriptive minimum on this sample, not statistical superiority or
deployment utility; rank uncertainty is unquantified.}
\label{tab:locked-cost}
\end{table*}

Together, these results show that editing the reserved output score is a
compact way to move the short-segment abstention operating point. They do
not establish a privileged causal locus, message-free-specific use of the
trained direction, improved risk--coverage over an external detector, or a
known training prior for stock checkpoints.

\section{RQ3: Boundary Tests Across Evaluated Systems}
\label{sec:class}

These developmental analyses test the boundary of the Whisper result; they
are not locked replications. We do not assume one mechanism across model
families. AED and translation models use EOT or EOS; CTC and RNN-T systems
use alignment blanks with different decoding behavior. Multilingual
results require multilingual or per-language calibration.

\subsection{Other AED and Translation Models}

Canary-1B and OWSM v3.1 are AED recognizers unrelated to Whisper; NLLB-200
\citep{nllb2022} and MarianMT \citep{junczys2018marian, tiedemann2020opusmt}
are encoder--decoder translators from different families. In each model,
the step-0 null-token margin distinguishes message-free from meaningful
input, and a scalar null-token bias reduces output on the evaluated
dose-response subset (supplementary
Table~\ref{tab:scalar-summary}).

The cost differs substantially. Canary has a low-cost measured point,
whereas OWSM truncates real speech as the bias increases. For OWSM,
$\beta=9$ is the empirical error-balance point and $\beta=10$ is the
lower-LOR endpoint. Canary changes LOR $79.3\to0.0\%$ with WER
$7.02\to6.94$ at its measured point, whereas OWSM changes
$90.7\to1.3\%$ with WER $8.19\to20.9$. The full grid is in supplementary
Table~\ref{tab:owsmgrid}; these are different operating points, not
conflicting estimates.

The translation models show a controllable EOS coordinate but not the same
operating-point prediction. Their message-free sources are empty strings,
whitespace, punctuation-only strings (e.g., ``\dots'' or ``!!!''), and
random-character strings. Increasing EOS bias reduces retained lexical
output on those sources while also reducing real-source length and
translation quality: at the reported points, NLLB lexical-output rate changes
$27.5\to1.0\%$ while COMET changes $83.4\to80.9$, and Marian's rate changes
$7.0\to0.0\%$ while COMET changes $79.9\to75.6$. Supplementary
Tables~\ref{tab:scalar-summary} and~\ref{tab:mtquality} report
the selected dose responses, BLEU, chrF, and COMET \citep{rei2020comet}.
SeamlessM4T and
Parakeet are stock checks only and are excluded from the intervention
claim.

\subsection{CTC and RNN-T Systems}

CTC and RNN-T blanks are natively unsuppressed, and their non-speech
outputs are usually short fragments rather than fluent text. Anchor
fine-tuning helps the two strongest CTC systems in the battery, while a
blank-bias scalar helps several others. Crossed interventions show that no
single method is uniformly safest (supplementary
Table~\ref{tab:blank}). These results concern blank-output control and
do not establish the AED mechanism for CTC or RNN-T.

\subsection{Multilingual Evaluation}

An English-trained row does not transfer reliably. A separate multilingual
row is calibrated on seven language tags and evaluated on eight held-out
tags, with Korean as the largest residual. The other model families use
one scalar per evaluated language. These are multilingual or per-language
calibrations, not zero-shot transfer. On held-out speech, large and turbo
retain median empty-output rates of $0.0\%$ and median CER changes of
$+1.7$ and $+0.1$ points while reducing mean LOR
$81.8\!\to\!8.3\%$ and $77.6\!\to\!5.8\%$. Smaller scales reach low
residual LOR at much more abstaining points (median empty output
$29.1$--$84.2\%$); Korean is the worst residual language at every scale.
Full lexical-output and speech-cost rows are in supplementary
Tables~\ref{tab:enpatch} and~\ref{tab:xling-cost}.

\subsection{Representation-Editing Baselines}

On the inspected Whisper-small training matrix, the 768-parameter EOT row is
the smallest update, reaching zero developmental LOR at the first checkpoint
without changing probe WER (Table~\ref{tab:cost}). Larger and randomly located
updates also reach zero, establishing parameter efficiency rather than safety
or a unique locus. AudioSAE and reduced LLT use different procedures and
provide no safer common operating point; their full results and
cross-architecture Calm-Whisper checks remain in supplementary
Tables~\ref{tab:audiosae} and~\ref{tab:calmxarch}.

\begin{table}[t]
\centering
\footnotesize
\setlength{\tabcolsep}{3pt}
\begin{tabular}{llrrr}
\toprule
Method & Locus & Params & 0\% HR at & WER \\
\midrule
\textbf{\texttt{eot\_row}} & \textbf{EOT row} & $\mathbf{+768}$ & \textbf{50} & \textbf{5.11} \\
LoRA ($r{=}8$)      & q/v proj.\   & 884{,}736    & 50        & 5.5 \\
decoder-only FT     & decoder      & 154M         & 50        & 5.6 \\
full FT             & all          & 242M         & 50        & 5.7 \\
random-matched      & random scal. & 95{,}232     & 300       & 5.2 \\
LayerNorm (all)     & LN           & 95{,}232     & 150       & 7.6 \\
Calm-Whisper        & 3 heads      & $+590{,}208$ & 50$^{+}$  & 5.78 \\
AudioSAE $\alpha=1$ & enc.\ SAE & $+113$M & no$^{a}$ & 3.74 \\
AudioSAE $\alpha=8$ & enc.\ SAE & $+113$M & $\alpha=8^{a}$ & 100.0 \\
LLT proxy           & layer-mix    & $+12$        & no$^{b}$ & 14.2 \\
\bottomrule
\end{tabular}
\caption{Cost comparison on Whisper small. Training rows report the first
zero-HR step and held-out probe WER (stock $5.11$); $50^{+}$ also requires
per-model head screening. $^{a}$The released AudioSAE is evaluated directly
on our $630$-clip battery and test-clean protocol (stock WER $3.73$):
$\alpha=1$ leaves HR at $94.6\%$, whereas the first zero-HR point,
$\alpha=8$, gives WER $100\%$ (full sweep in supplementary
Table~\ref{tab:audiosae}). $^{b}$The reduced LLT layer-mixing proxy
raises HR $92\to100\%$ and is not a full teacher-distillation reproduction.}
\label{tab:cost}
\end{table}

\section{Limitations and Scope}
\label{sec:limits}

We measure and perturb normalized lexical output on whole-segment empty
targets, where every retained lexical token is fabricated by construction.
Interleaved fabrication is excluded: it requires per-token decisions on
non-empty targets, has no label-free detector, and global abstention can
harm it.

Lexical-output suppression and false silence require joint evaluation. Our
common-cost comparison reports whether an operating point empties at most
$10\%$ of real-speech clips, an illustrative cap rather than application
utility. Within it, external state and VAD gates achieve lower developmental
LOR (Table~\ref{tab:common-cost-main}). A row packages a state-dependent score
change in one checkpoint coordinate, a format rather than utility advantage.
The locked $C_\kappa$ family reverses the point-estimate ranking under the
displayed weights; cost-gap uncertainty is unquantified.

The model-dependent safety ledger (supplementary
Tables~\ref{tab:ledger}--\ref{tab:earnings}) shows tiny and base
improving on degraded speech through insertion and substitution reductions
despite more deletions, whereas four stronger Whisper scales increase WER
and spontaneous-speech costs vary sharply. Medium and OWSM retain narrow
low-cost regions. In long-form decoding, the row reduces gap words but raises
WER; correctly implemented VAD segmentation dominates both axes. Thus
evidence for checkpoint editing is restricted to short, empty-target segments
and does not establish superiority over a standard front-end on locked data.

Analytic calibration and evaluation pools are disjoint, but clean-speech
constraints do not certify new acoustics. Deployment needs target-domain
calibration, a non-speech base-rate estimate, and explicit false-silence cost.
Inspected developmental data include stress battery and reduced dose-response
subsets. The source-disjoint, collision-audited locked 300/300 cost evaluation
includes previously reported test-clean, lacks the intended offline
spontaneous stratum, and is not a complete deployment simulation. Missing
anonymous artifact locator and scale-specific batch-size/learning-rate
settings leave the author-reported freeze and exact runs independently
unauditable from PDFs.

No human transcript adjudication supports the headline endpoint. For every
evaluated message-free span, an empty lexical reference makes any output
remaining after documented non-lexical event-tag removal fabricated by
construction, consistent with established non-speech protocols
\citep{baranski2025investigation,calmwhisper2025}. Speech-containing inputs,
where hallucination and mistranscription require adjudication, remain
excluded.

Current normalization leaves 13 of 15 models invariant, moving Whisper
medium $0.31$ and OWSM $6.67$ points by hiding tag-only, number-only, and
wrong-script outputs. The incomplete blinded protocol provides no positive
evidence. False silence, the primary foreseeable harm, peaks in locked low-SNR
and accented groups. LLT remains a reduced proxy;
beam search untested. Avoid small-model use on degraded or spontaneous speech,
long-form audio, or when false silence is costlier than fabrication.

\section{Conclusion}

Developmentally, decoder states and native null-token scores separate the
evaluated conditions, but clip-stratified probes do not establish a
source-independent message-absence representation. Scalar bias, analytic row
graft, and trained-row edits are parameter-compact diagnostic instruments for
moving the short-segment operating point, without establishing a privileged
mechanism or safer risk frontier.

On the only post-freeze sample, Whisper-small's trained row moves LOR
$91.7\%\to3.3\%$ and deleted reference words/min $1.58\to33.05$. It never
minimizes the displayed $C_\kappa$ grid: bias $b=5$ has the lowest point
estimate for $\kappa=0.5,1$, and stock for $\kappa\ge2$. This descriptive
five-condition ranking has unquantified rank uncertainty and no locked
comparison with stronger developmental gates.

Results are consistent with abstention miscalibration under some evaluated
conditions. Learned margins' positive, heterogeneous training-log-odds slopes
do not establish the classical unit coefficient; failed translation
operating-point predictions limit prior-shift explanations. CTC/RNN-T,
translation, and multilingual results require family-specific interpretations.

Developmental cross-family, translation, and multilingual boundaries carry
model-dependent safety costs; VAD is the evaluated long-form preference.
Null-coordinate editing diagnoses fabrication--deletion tradeoff, not
safe abstention. Interleaved hallucination, beam search, deployment utility,
and zero-shot transfer remain excluded.

% References and End of Main Paper
\small
\bibliography{references}
\normalsize

% ======================================================================
% Supplement (as Appendix) -- starts on its own page; single column so
% wide tables no longer need the table* float mechanism.
% ======================================================================
\clearpage
\onecolumn
\appendix

\section{Extended Validation and Audit Trail}
\label{sec:extended-validation}

This section reports the post-freeze validation summarized in the main
paper. It adds a source-group manifest, a data-role ledger, an untouched
source evaluation, one shared cost definition, and seed replication. The
values below were generated from per-example result files rather than
transcribed into the analysis scripts.

\FloatBarrier

\subsection{Data Roles and Source Disjointness}

The role ledger contains 12 pools assigned to four roles: fit/calibration,
selection, developmental reporting, and locked reporting. The developmental
battery contains 1060 clips derived from 869 source recordings. Resolving
ESC-50 and UrbanSound8K identifiers back to source recordings reveals 21
UrbanSound8K clip pairs with overlapping spans; these pairs stay within a
single role. The locked set contains 300 non-speech and 300 speech clips.
Disjointness is checked by sample SHA-1, source key, and speaker. As a
falsification test, inserting a calibration clip makes the checker refuse
to run.

Locked non-speech comprises silence from an unused random seed plus MUSAN,
ESC-50, UrbanSound8K, and music recordings absent from all fit and selection
pools. Locked speech comprises LibriSpeech test-clean, WHAM!-degraded and
low-SNR mixtures, and accented FLEURS. The planned spontaneous stratum was
unavailable offline and is marked absent. Test-clean had appeared earlier
as a reported benchmark, so it is disjoint from fitting and selection but
not previously unseen.

\begin{table}[htbp]
\centering
\scriptsize
\setlength{\tabcolsep}{3pt}
\begin{tabular}{p{2.0cm}p{3.0cm}p{3.0cm}p{3.2cm}p{1.1cm}}
\toprule
Result & Fit/calibration pool & Selection rule or pool & Reporting pool & Inspected? \\
\midrule
Frozen-state probe &
Full battery within five clip folds &
Cross-validated probe; PCA fitted within each analysis &
The same 630 non-speech and 430 speech clips & yes \\
Trained EOT row &
MUSAN calibration half, synthetic silence, LibriSpeech dev-clean &
Common step 50; first-zero rows summarize trajectories &
Full 630-clip non-speech battery and named speech subsets & yes \\
Analytic row graft &
Disjoint MUSAN half and LibriSpeech dev-clean speakers &
Admissible interval on the calibration pool &
Stress battery and test-clean, disjoint from fitting & yes \\
AED/NMT scalar bias &
No parameter fit; margins logged on fixed dose-response data &
Bias chosen on 150 non-speech and 60 speech clips &
The same dose-response subset; MT quality on WMT19 & yes \\
Common-cost comparison &
Each method's stated calibration &
Lowest LOR satisfying the illustrative $10\%$ speech-side cap &
The same 630 non-speech and 430 speech clips & yes \\
Multilingual row/scalars &
Seven Whisper tags or named language calibration splits &
Frozen row/scalars after calibration &
Eight held-out tags or held-out language clips & yes \\
Long-form analysis &
Previously calibrated row or fixed VAD configuration &
No selection on the reporting clips &
40 clips with speech and message-free gaps & yes \\
Locked $C_\kappa$ &
All methods and costs frozen before decoding &
No selection on locked outputs &
Source-disjoint 300 non-speech / 300 speech clips & no \\
\bottomrule
\end{tabular}
\caption{Fit, selection, and reporting roles. ``Inspected'' means that
reporting-pool outputs were available during method development. The locked
cost ledger entry is the only post-freeze reporting pool; exact source IDs
and computed collisions are author-reported as stored in project manifests
that are not linked from the review PDF.}
\label{tab:decision-record}
\end{table}
\FloatBarrier

\FloatBarrier

\subsection{Matched Developmental Cap Comparison}

All arms in Table~\ref{tab:common-cost} use the same 630 non-speech and 430
speech clips. We mark whether a method is within the illustrative cap of
rejecting or emptying no more than $10\%$ of real-speech clips. The first
four rows are separately thresholded detectors. Their exact achieved
speech-side rates are absent from the available summary, which preserves
only that the selected thresholds are within the cap. The last four rows
are decoder/checkpoint conditions.

\begin{table}[htbp]
\centering
\scriptsize
\setlength{\tabcolsep}{3pt}
\begin{tabular}{lrrc}
\toprule
Method & LOR & speech-side rate & within cap \\
\midrule
native EOT margin & 89.8 & $\le10.0$ & yes \\
native \texttt{no\_speech\_prob} (SOT) & 4.3 & $\le10.0$ & yes \\
logistic gate (frozen state) & 0.0 & $\le10.0$ & yes \\
FireRedVAD front-end & 1.3 & $\le10.0$ & yes \\
\midrule
stock & 96.2 & 0.0 & yes \\
constant EOT bias $b=5$ & 24.3 & 7.4 & yes \\
trained \texttt{eot\_row} & 0.2 & 67.7 & no \\
analytic graft $\lambda=2$ & 0.0 & 35.6 & no \\
\bottomrule
\end{tabular}
\caption{Matched, inspected Whisper-small comparison (percent). LOR is the
normalized lexical-output rate. Speech-side rate is rejection for detectors
and empty output for checkpoint interventions. The external detectors have
lower developmental LOR while satisfying the coarse cap; this is not a
locked efficacy or superiority comparison.}
\label{tab:common-cost}
\end{table}

\begin{table}[htbp]
\centering
\scriptsize
\setlength{\tabcolsep}{3.5pt}
\begin{tabular}{lccc ccccc}
\toprule
 & \multicolumn{3}{c}{trained EOT row}
 & \multicolumn{5}{c}{analytic graft} \\
\cmidrule(lr){2-4}\cmidrule(lr){5-9}
Model & stock LOR & patched LOR & patched WER (stock)
 & $\lambda^\star$ & $[\lambda_{\mathrm{lo}},\lambda_{\mathrm{hi}}]$ & in?\
 & LOR & test-clean WER \\
\midrule
tiny     & 94.9 & 0.0 & \textbf{10.61 (10.61)} & 1.017 & $[0.45, 2.28]$   & yes & 0.3 & $7.54\to8.22$ \\
base     & 99.7 & 0.0 & \textbf{7.74 (7.74)}   & 0.507 & $[0.51, \infty)$ & yes & 2.5 & $5.04\to4.84$ \\
small    & 96.2 & 0.0 & \textbf{5.11 (5.11)}   & 2.406 & $[0.53, 10.91]$  & yes & 0.5 & $3.44\to4.16$ \\
medium   & 86.8 & 0.0 & \textbf{3.59 (3.59)}   & 0.497 & $[0.50, \infty)$ & yes & 1.3 & $2.88\to7.60$ \\
turbo    & 74.3 & 0.0 & 3.59 (2.69)   & 0.661 & $[0.66, \infty)$ & yes & 3.5 & $2.16\to2.26$ \\
large-v3 & 77.0 & 0.0 & 2.47 (2.41)   & 0.824 & $[0.82, \infty)$ & yes & 8.4 & $2.02\to2.08$ \\
\bottomrule
\end{tabular}
\caption{Developmental trained-row and analytic-graft summary by Whisper
scale. Trained-row LOR is on the inspected battery at common step 50; WER
is on the held-out dev-clean probe and parentheses give stock WER. The
analytic graft uses disjoint calibration pools and reports battery LOR plus
test-clean WER. Values and interval bounds are rounded independently.}
\label{tab:fix-summary}
\end{table}

\FloatBarrier

\subsection{Locked Fabrication--Deletion Sensitivity}

The primary locked endpoint is
\[
C_\kappa=F+\kappa D,
\]
where $F$ is total words retained by the specified lexical normalizer
divided by total non-speech audio minutes and $D$ is total reference-word
deletions divided by total speech audio minutes. The effective weight
$\kappa$ combines relative
exposure/prevalence and per-word harm; this is a conditional-rate
sensitivity index, not expected deployment utility or total WER. Methods,
configurations, and $\kappa\in\{0.5,1,2,5,10\}$ were pre-specified and
author-frozen before the 300/300 decode.

\begin{table}[htbp]
\centering
\scriptsize
\setlength{\tabcolsep}{3pt}
\begin{tabular}{lrrrrrrr}
\toprule
Method & $F$ & $D$ & $C_{0.5}$ & $C_1$ & $C_2$ & $C_5$ & $C_{10}$ \\
\midrule
stock & 32.5 & 1.6 & 33.3 & 34.1 & \textbf{35.7} & \textbf{40.4} & \textbf{48.3} \\
trained EOT row & 4.3 & 33.0 & 20.8 & 37.3 & 70.4 & 169.5 & 334.8 \\
constant EOT bias $b=5$ & 10.1 & 17.6 & \textbf{18.9} & \textbf{27.7} & 45.3 & 98.0 & 185.7 \\
constant EOT bias $b=6$ & 6.4 & 34.6 & 23.7 & 41.0 & 75.8 & 179.9 & 353.6 \\
analytic graft $\lambda=2$ & 35.4 & 17.4 & 44.1 & 52.8 & 70.1 & 122.1 & 208.8 \\
\bottomrule
\end{tabular}
\caption{Locked $C_\kappa$ (lower is better). $F$ and $D$ are rounded to
one decimal; costs use unrounded components. Boldface marks the descriptive
minimum on this 600-clip sample. Rank uncertainty is unquantified, the
weights are illustrative, and external gates were not run on this set.}
\label{tab:locked-cost-full}
\end{table}

The trained row reduces clip-level LOR from $91.7\%$ to $3.3\%$
(zero on silence, music, events, and urban audio), but deletes $33.05$
reference words/min versus stock's $1.58$. Its groupwise false-silence rate
is $8.0\%$ on clean speech, $42.7\%$ on low-SNR speech, and $38.7\%$ on
accented speech. These worst groups are reported rather than averaged away.
They characterize the trained row only; groupwise rates for every locked
condition are not available in the displayed result record.

\FloatBarrier

\subsection{Held-Out Multilingual Speech Cost}

For each language, 40 clips calibrate the multilingual row and 60 disjoint
clips report hallucination, real-speech CER, and empty output.

\begin{table}[htbp]
\centering
\scriptsize
\setlength{\tabcolsep}{4pt}
\begin{tabular}{lrrrrr}
\toprule
Model & stock LOR & fixed LOR & worst-language LOR & median empty & median $\Delta$CER \\
\midrule
large-v3 & 81.8 & 8.3 & pt: 20.0 & 0.0 & 1.7 \\
turbo    & 77.6 & 5.8 & es: 11.7 & 0.0 & 0.1 \\
base     & 92.3 & 3.1 & ko: 46.7 & 29.1 & 24.7 \\
small    & 79.0 & 2.0 & ko: 8.3  & 47.5 & 46.4 \\
tiny     & 71.8 & 3.8 & ko: 56.7 & 72.5 & 56.7 \\
medium   & 75.0 & 0.3 & ko: 3.3  & 84.2 & 84.2 \\
\bottomrule
\end{tabular}
\caption{Held-out 15-language Whisper evaluation (percent). Large-v3 and
turbo retain real speech at the calibrated point; smaller scales reach low
mean LOR by abstaining heavily. Korean is the worst residual language at all
six scales.}
\label{tab:xling-cost}
\end{table}

\begin{table}[htbp]
\centering
\scriptsize
\setlength{\tabcolsep}{4pt}
\begin{tabular}{llcclccc}
\toprule
Model & Task & AUROC$_0$ & lexical LOR & Speech/MT change & $\beta$
 & $\beta^\star$ pred.\ & $\beta^\star$ emp. \\
\midrule
Canary-1B & ASR & $1.000$ & $\mathbf{79.3\to0.0}$ & WER $\mathbf{7.02\to6.94}$ & $8$  & \textbf{8.0}  & \textbf{8.0} \\
OWSM v3.1 & ASR & $0.998$ & $90.7\to1.3$ & WER $8.19\to20.9$ & $10$ & 10.0 & 9.0 \\
NLLB-200  & NMT & $0.999$ & $27.5\to1.0$ & COMET $83.4\to80.9$ & $6$ & 3.5  & 6.0 \\
MarianMT  & NMT & $0.911$ & $7.0\to0.0$  & COMET $79.9\to75.6$ & $6$ & 9.5  & 3.0 \\
\bottomrule
\end{tabular}
\caption{Developmental scalar null-token-bias summary (percent except
$\beta$). AED dose response uses 150 non-speech and 60 speech clips; NMT
quality uses real-source WMT19. $\beta^\star$ is selected and reported on
the same grid, so these are descriptive rather than locked estimates.
COMET is multiplied by 100.}
\label{tab:scalar-summary}
\end{table}
\FloatBarrier

\FloatBarrier

\subsection{Conditional Prior-Shift Derivation}
\label{sec:prior-shift}

Let $z=1$ mark message-free input (abstain), $z=0$ input carrying a message
(emit), and $\pi=P(z{=}1)$. The decoder's step-0 score is the abstain margin
$m(x)$. Changes in acoustics, language tags, or decoder context violate the
fixed-class-conditional premise below.

\begin{proposition}[Calibrated-margin prior correction]
Assume label shift with fixed $p(x\mid z)$ and a unit-scale calibrated
training margin,
\[
m_{\mathrm{train}}(x)=
\log\frac{p(x\mid z=1)}{p(x\mid z=0)}
+\mathrm{logit}(\pi_{\mathrm{train}}).
\]
Then
$m_{\mathrm{test}}(x)=m_{\mathrm{train}}(x)
+\mathrm{logit}(\pi_{\mathrm{test}})
-\mathrm{logit}(\pi_{\mathrm{train}})$; for our margin, the constant can be
added to the EOT logit.
\end{proposition}

This is the classical Bayes identity
$\mathrm{logit}\,P(z{=}1\mid x)=\log\mathrm{LR}(x)+\mathrm{logit}(\pi)$
\citep{saerens2002,elkan2001foundations,lipton2018bbse}. It is not a theorem
about an arbitrary neural EOT-versus-content margin. If the neural margin
approximates the assumed score, three diagnostics follow: a row trained at
non-speech fraction $p$ should install a margin linear in
$\mathrm{logit}(p)$ with unit slope; a deployment scalar is the classical
prior correction; and, under an additional one-step symmetric-margin and
equal-cost approximation, an error-balance bias can be predicted from
step-0 margins. The first and third are applicability checks, not evidence
that prior shift caused the stock behavior.

\begin{figure}[htbp]
\centering
\includegraphics[width=0.90\columnwidth]{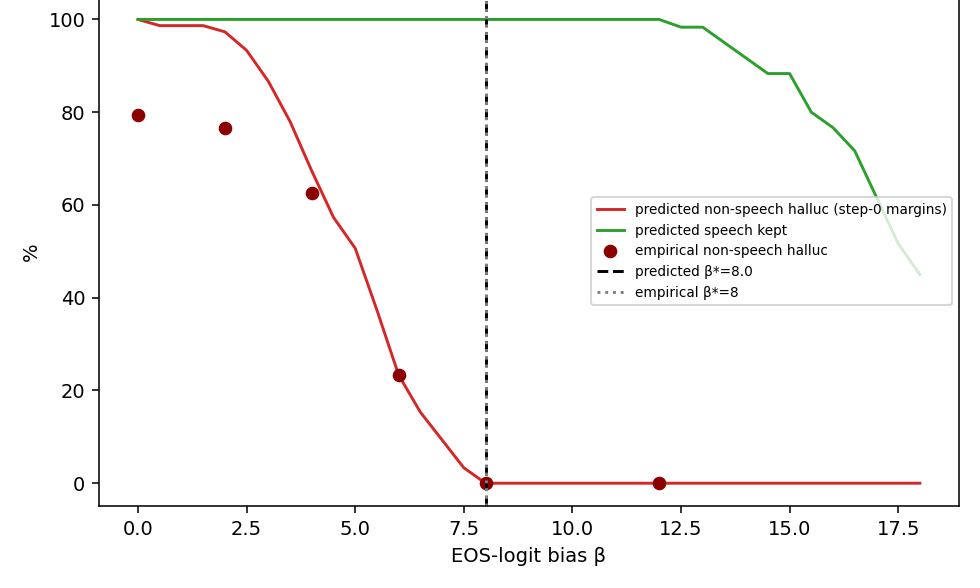}
\caption{Developmental Canary-1B dose response on 150 non-speech and 60
speech clips. Curves are predicted from step-0 margins; points are measured
decodes. The dashed line marks the descriptive error-balance grid point,
where missed and false abstentions are closest.}
\label{fig:betastar}
\end{figure}

\begin{table}[htbp]
\centering
\footnotesize
\begin{tabular}{lccc}
\toprule
Model & slope & $95\%$ CI & $r$ \\
\midrule
tiny     & 0.96 & $[0.41,1.50]$ & 0.90 \\
base     & 0.80 & $[0.52,1.08]$ & 0.96 \\
small    & 1.05 & $[0.65,1.45]$ & 0.95 \\
medium   & 1.53 & $[1.01,2.06]$ & 0.96 \\
turbo    & 1.89 & $[1.25,2.52]$ & 0.96 \\
large-v3 & 1.69 & $[1.18,2.21]$ & 0.97 \\
\midrule
pooled RE & 1.29 & $[0.94,1.65]$ & n.a. \\
\bottomrule
\end{tabular}
\caption{Developmental replicated analysis of learned abstention margin
versus $\mathrm{logit}(p)$ over seven training fractions per scale. These
run-level intervals summarize the evaluated trainings; heterogeneity is
high and equivalence to unit slope is not established.}
\label{tab:baserate-replicated}
\end{table}

\FloatBarrier

\subsection{Statistical, Scoring, and Protocol Audits}

The six base-rate slopes are combined with a random-effects model because
heterogeneity is high ($\tau^2=0.163$, $I^2=84\%$). The pooled slope is
$1.29$ with $95\%$ CI $[0.94,1.65]$; the scale trend is $+0.214$ per scale
step ($t=1.99$, 4 df, $p=0.118$). No per-scale departure from unity
survives Holm or Benjamini--Hochberg correction, but equivalence to unity
under $[0.8,1.2]$ also fails. Seed replication uses 18 independent
resamples with five seeds on small/medium/large and three on
tiny/base/turbo, giving 126 row trainings. Batch-size recipe sensitivity is
logged separately from seed variation. Separately, the reported Wilson
intervals and McNemar tests use clips as units on an inspected battery.
They are finite-battery summaries, are not clustered by source recording,
and do not quantify corpus-transfer or method-selection uncertainty. The
PDF-level record does not establish whether the 21 overlapping-span
UrbanSound pairs share cross-validation folds.

Normalization sensitivity is zero on 13 of 15 cached models, $0.31$
percentage points on Whisper medium, and $6.67$ points on OWSM. For OWSM,
the current normalizer hides 34 tag-only, 8 number-only, and 7 wrong-script
clips. This is why the main paper treats lexical normalization as a
measurement boundary rather than semantic adjudication.

The audit infrastructure includes a blinded 150-clip sheet with opaque arm
IDs, shuffled rows, six dimensions for two annotators, and an adjudication
column. The sheet is prepared; annotation has not been completed. For
multilingual NMT, a point is marked within cap only if at least $90\%$ of
genuine-source translations remain non-empty. NLLB Chinese and Marian
Spanish have no point within that cap and are marked as failures rather than
included in the mean.

\FloatBarrier

\section{Implementation and Procedural Checks}
\label{sec:pipeline}

\paragraph{Common training recipe.}
Training uses an example-balanced 50/50 anchor mix: MUSAN-training-half
noise plus synthetic silence with an empty target, and lower-cased
LibriSpeech dev-clean speech with its transcript. Runs use weight decay
$0$, gradient clipping at $1.0$, 400 steps, and evaluation every 50 steps.
AdamW is used below 50M trainable parameters and Adafactor above; the EOT
row uses learning rate $10^{-3}$. Decoder inputs are passed explicitly and
all stock/patch evaluations use greedy temperature-0 decoding with
\texttt{begin\_suppress\_tokens=None}. The reported step-50 row is fixed
across scales; ``first zero'' is adaptive and descriptive. Scale-specific
batch sizes and the broader learning-rate grid remain in the prepared
configuration inventory and are not independently recoverable from this
PDF alone.

Four HuggingFace-pipeline defaults each silently reproduce ``the model
resists fine-tuning'' without any involvement of the model. We list the
symptom, the diagnostic, and the fix.

\paragraph{Prefix masking.} Masking the forced prefix in \texttt{labels} with
$-100$ while letting the collator rebuild \texttt{decoder\_input\_ids} via
\texttt{shift\_tokens\_right} turns every $-100$ into the pad token (EOT), so
the decoder trains on the garbage context [\texttt{sot}, \texttt{eot},
\texttt{eot}, \dots]. Diagnostic: teacher-forced loss near $\ln V \approx
10.7$. Fix: pass \texttt{decoder\_input\_ids} explicitly.

\paragraph{Begin-suppress ban.} \texttt{begin\_suppress\_tokens} forbids EOT
as the first token, so a model that has perfectly learned
$P(\mathrm{EOT}\mid\text{noise})\approx 1$ still emits content; on tiny,
LayerNorm tuning evaluated with the ban enabled pushed hallucination from
$89\%$ to $100\%$. Fix: evaluate fine-tuned models with
\texttt{begin\_suppress\_tokens=None} (verified a no-op on stock models).

\paragraph{Token-averaged loss.} A non-speech example contributes one target
token (EOT) against roughly forty for a speech example, so token averaging
gives the anti-hallucination gradient $\sim\!2\%$ of the weight;
$-\log P(\mathrm{eot})$ barely moves ($7.3 \to 6.4$ over 400 steps). Fix:
example-balanced loss (mean over per-example mean token cross-entropy).

\paragraph{Reentrant checkpointing.} With \texttt{use\_reentrant=True} (the
default), parameters inside checkpointed blocks whose inputs do not require
grad receive no gradient: on turbo only the two final LayerNorms out of 156
tensors actually trained. Fix: \texttt{use\_reentrant=False}.

\paragraph{Selection and scoring checks.} Checkpoint selection by minimum
clean-probe WER is blind to over-abstention and prefers late, destructive
checkpoints (up to $62\%$ of spontaneous out-of-domain speech left empty;
Table~\ref{tab:earnings}); selection must include out-of-domain speech. And
scorers must strip non-speech tags (e.g.\ \texttt{[music]}) before counting
fabricated letters, or hallucination is over-counted on models that emit
tags.

\paragraph{The sixteen regimes.} full; decoder-only; encoder-only; LoRA
($r{=}8$, q/v); final decoder layer; LayerNorm all / $\gamma$ only / $\beta$
only / encoder-only / decoder-only / final pair; BitFit (all biases except
LayerNorm); \texttt{eot\_row} (the EOT unembedding row); random-matched
(random scalars via gradient mask, exactly the LayerNorm count, LayerNorm
excluded); random-colocated (random scalars in the attention/MLP tensors of
the same blocks); calm-heads (top-3 decoder self-attention heads by ablation
screening).

\FloatBarrier

\section{Analytic Row-Graft Procedure}
\label{sec:lambda}

The analytic graft is a supervised linear recalibrator of one output
coordinate. It is not a scalar prior-odds correction. The following
pseudocode records the reported construction.

\begin{enumerate}
\item Collect frozen decoder states from labeled calibration data. Positive
states are non-speech at the first output position and genuine
end-of-utterance positions. Negative states are mid-utterance speech
positions.
\item Fit a logistic direction $w$ to those states.
\item For each non-speech clip, convert the requirement that EOT beat every
content logit into a lower bound on $\lambda$. Set
$\lambda_{\mathrm{lo}}$ to the largest lower bound.
\item For each speech clip and mid-utterance position, convert the
requirement that EOT remain below the correct content decision into an
upper bound. Set $\lambda_{\mathrm{hi}}$ to the smallest upper bound.
\item Reject the graft if
$\lambda_{\mathrm{lo}}>\lambda_{\mathrm{hi}}$. Otherwise choose
$\lambda^\star=\sqrt{\lambda_{\mathrm{lo}}\lambda_{\mathrm{hi}}}$ when
the upper bound is finite and use $\lambda_{\mathrm{lo}}$ when it is
unbounded.
\item Apply
$w_{\mathrm{EOT}}\leftarrow w_{\mathrm{EOT}}+\lambda^\star w$ and evaluate
on the designated reporting pool.
\end{enumerate}

The interval uses only the calibration pool. Because its speech
constraints are clean, feasibility does not certify degraded or
out-of-domain speech.

\FloatBarrier

\section{Reporting and Reproducibility Details}
\label{sec:decision-record}

The fit, selection, and reporting pools are recorded in
Table~\ref{tab:decision-record}. This section fixes
the terminology used across the detailed result tables and lists the
contents of the reproducibility archive.

\paragraph{Subset-dependent terminology.}
\texttt{eot\_row@50} always denotes the common step-50 trained-row
checkpoint evaluated on the full battery. ``First zero'' denotes the first
inspected checkpoint reaching zero hallucination in a training trajectory
and is used only as an adaptive descriptive summary. For OWSM,
$\beta=9$ is the empirical error-balance point, while $\beta=10$ is the
lower-hallucination point. Every WER is interpreted only with the
denominator and source group named in its table.

\paragraph{Artifact inventory and release status.}
The author-reported prepared artifact inventory contains immutable model identifiers and
revisions, environment lockfiles, source-level split manifests and
checksums, normalization and tag-removal code, random seeds, training and
decoding configurations, checkpoint and threshold selection logs, learned
rows and biases, per-example stock and patched outputs, and one
reproduction command per table. It resolves short labels including
wav2vec2-large, HuBERT-large, WavLM-100h, wav2vec2-base, Conformer-CTC,
Emformer-RNN-T, and Conformer-RNN-T. A stable anonymous locator and
snapshot hash are not included in the current review bundle, so public
availability is not asserted here. The authors intend to release the
artifacts upon publication, subject to upstream model and dataset licenses.
Consequently, the freeze, exact revisions, scale-specific batch sizes,
broader learning-rate grid, and one-command table reconstruction cannot be
independently checked from the review PDFs.

\paragraph{Data use and foreseeable harm.}
The experiments use cited public audio and translation corpora; source
audio remains governed by its original license, and the release inventory
redistributes only material permitted by those terms. No new human-subject
data were collected. The principal foreseeable harm is false silence:
the locked row suppresses $42.7\%$ of low-SNR and $38.7\%$ of accented
speech in the measured groups. These disparities motivate the explicit
non-deployment recommendation for the row at that operating point.

\FloatBarrier

\section{Deployment Checklist}
\label{sec:deployment-checklist}

\begin{enumerate}
\item Build a target-domain calibration set containing message-free,
clean-speech, and degraded-speech inputs, with no overlap with final
evaluation.
\item Estimate the expected message-free base rate only if the scalar
prior-shift model is appropriate; otherwise select a point from a labeled
cost curve.
\item Specify the relative cost of fabricated output and false silence,
then report both at the selected point.
\item Verify that the decoder permits the null token at the intended output
position and that stock and patched configurations are identical.
\item Recalibrate after changes in acoustic domain, language tag, decoder
policy, segmentation, or long-form context. Do not extrapolate the
single-segment results to interleaved hallucination.
\end{enumerate}

\FloatBarrier

\section{Stress Set, Statistics, and Spontaneous Speech}

\begin{table}[htbp]
\centering
\footnotesize
\begin{tabular}{llr}
\toprule
Category & Source & $n$ \\
\midrule
silence\_pure     & synthetic zero $+$ dither   & 40   \\
silence\_room     & pink noise (room tone)      & 40   \\
noise\_stationary & MUSAN noise                 & 150  \\
noise\_events     & ESC-50                      & 150  \\
noise\_urban      & UrbanSound8K                & 150  \\
music             & MUSAN music (instrumental)  & 100  \\
speech\_speed     & FLEURS, time-stretched      & 150  \\
overlap\_2/3spk   & FLEURS speaker mixes        & 100  \\
speech\_noise     & FLEURS $+$ MUSAN            & 120  \\
speech\_clean     & FLEURS \texttt{en\_us}      & 60   \\
\midrule
Total             &                             & 1060 \\
\bottomrule
\end{tabular}
\caption{The 1060-clip stress battery (3.1 h, 16 kHz mono, clips $\le 30$ s;
seed 20260611). Non-speech strata carry an empty reference; speech strata are
FLEURS-derived. Exclusions: UrbanSound8K classes that may contain words
(children\_playing, street\_music) and vocal MUSAN tracks are dropped. Speech
sources are RMS-normalized to $-20$ dBFS.}
\label{tab:stress}
\end{table}

\begin{table}[htbp]
\centering
\footnotesize
\begin{tabular}{llcccccccc}
\toprule
& & test-clean & \multicolumn{7}{c}{Lexical-output rate (\%) by stratum} \\
\cmidrule(lr){4-10}
Model & Type & WER \% & silence & room & noise & events & urban & music & all \\
\midrule
Whisper tiny     & AED   & 7.54 & 100 & 100 & 86  & 96  & 99  & 97  & 94.9 \\
Whisper base     & AED   & 5.04 & 100 & 100 & 100 & 99  & 100 & 100 & 99.7 \\
Whisper small    & AED   & 3.44 & 100 & 50  & 97  & 100 & 100 & 100 & 96.2 \\
Whisper medium   & AED   & 2.88 & 0   & 0   & 99  & 100 & 100 & 98  & 86.8 \\
Whisper turbo    & AED   & 2.16 & 25  & 42  & 82  & 67  & 83  & 94  & 74.3 \\
Whisper large-v3 & AED   & 2.02 & 25  & 25  & 88  & 60  & 97  & 97  & 77.0 \\
Canary-1B        & AED   & 1.44 & 100 & 100 & 74  & 89  & 97  & 19  & 77.5 \\
OWSM v3.1        & AED   & 2.42 & 78  & 100 & 91  & 92  & 93  & 99  & 92.7 \\
\midrule
wav2vec2-large   & CTC   & 1.73 & 0   & 0   & 19  & 23  & 4   & 21  & 14.4 \\
wav2vec2-base    & CTC   & 3.26 & 0   & 0   & 33  & 37  & 13  & 69  & 30.8 \\
HuBERT-large     & CTC   & 1.97 & 0   & 0   & 75  & 82  & 35  & 100 & 61.4 \\
WavLM-100h       & CTC   & 6.57 & 0   & 0   & 36  & 47  & 20  & 74  & 36.2 \\
Conformer-CTC    & CTC   & 2.08 & 0   & 0   & 15  & 15  & 7   & 13  & 11.1 \\
\midrule
Emformer-RNN-T   & RNN-T & 4.43 & 0   & 0   & 21  & 23  & 5   & 83  & 24.8 \\
Conformer-RNN-T  & RNN-T & 1.65 & 0   & 0   & 5   & 3   & 0   & 0   & 1.9  \\
\bottomrule
\end{tabular}
\caption{Stock behavior of the fifteen-model core zoo. LibriSpeech
test-clean WER (Canary-1B and OWSM v3.1 on $500$ utterances, others on
$2620$) validates the pipeline. Lexical-output rate is the share of
non-speech clips with at least one retained letter after normalization.
Exact repository revisions are
author-reported as listed in a prepared model manifest that is not linked
from the review PDF.}
\label{tab:zoo}
\end{table}

\begin{table}[htbp]
\centering
\scriptsize
\setlength{\tabcolsep}{3pt}
\begin{tabular}{lll}
\toprule
Use & Evaluation set & Reason \\
\midrule
zoo/stats & full $630$ NS $+$ $430$ speech & headline rates/CIs \\
deployable ROC & fixed $300$ NS $+$ $430$ speech & threshold sweeps \\
AED/NMT bias & fixed $150$ NS $+$ $60$ speech & dose response \\
degraded ledger & fixed ID degradations & error alignment \\
spontaneous cost & Earnings-22/MUSAN-speech & external domain \\
long form & $40$ clips, $74$ min & chunk boundaries \\
\bottomrule
\end{tabular}
\caption{Evaluation-subset map. Subsets differ because large-model threshold
sweeps are substantially more expensive than one-pass headline evaluation.
Every intervention is compared with stock re-decoded on the same subset;
rates are not compared across rows. The first three subsets are fixed and
nested only where their listed counts coincide. NS denotes non-speech.}
\label{tab:subsets}
\end{table}

\begin{table}[htbp]
\centering
\footnotesize
\setlength{\tabcolsep}{3pt}
\begin{tabular}{lcccccc}
\toprule
& \multicolumn{2}{c}{stock} & \multicolumn{2}{c}{\texttt{eot\_row}@50}
& \multicolumn{2}{c}{analytic} \\
\cmidrule(lr){2-3}\cmidrule(lr){4-5}\cmidrule(lr){6-7}
Model & LOR & $95\%$ CI & LOR & $95\%$ CI & LOR & $95\%$ CI \\
\midrule
tiny     & 94.9 & $[92.9,96.4]$ & 0.0 & $[0.0,0.6]$ & 0.3 & $[0.1,1.2]$ \\
base     & 99.7 & $[98.8,99.9]$ & 0.0 & $[0.0,0.6]$ & 2.5 & $[1.6,4.1]$ \\
small    & 96.2 & $[94.4,97.4]$ & 0.0 & $[0.0,0.6]$ & 0.5 & $[0.2,1.4]$ \\
medium   & 86.8 & $[84.0,89.2]$ & 0.0 & $[0.0,0.6]$ & 1.3 & $[0.6,2.5]$ \\
turbo    & 74.3 & $[70.7,77.5]$ & 0.0 & $[0.0,0.6]$ & 3.5 & $[2.3,5.2]$ \\
large-v3 & 77.0 & $[73.5,80.1]$ & 0.0 & $[0.0,0.6]$ & 8.4 & $[6.5,10.8]$ \\
\bottomrule
\end{tabular}
\caption{Normalized lexical-output rates (LOR, \%; $n{=}630$ non-speech
clips) with clip-level Wilson $95\%$ intervals
\citep{wilson1927probable}. These conditional finite-battery summaries are
not source-clustered or corpus-transfer intervals. The trained row is
evaluated at common step 50; paired tests are in Table~\ref{tab:mcnemar}.}
\label{tab:stats}
\end{table}

\begin{table}[htbp]
\centering
\footnotesize
\setlength{\tabcolsep}{6pt}
\begin{tabular}{lrrrrr}
\toprule
Model & $b$ & $c$ & exact $p$ & Holm $p$ & OR \\
\midrule
tiny     & 598 & 0 & $2{\times}10^{-180}$ & $8{\times}10^{-180}$ & 1197 \\
base     & 628 & 0 & $2{\times}10^{-189}$ & $1{\times}10^{-188}$ & 1257 \\
small    & 606 & 0 & $8{\times}10^{-183}$ & $4{\times}10^{-182}$ & 1213 \\
medium   & 547 & 0 & $4{\times}10^{-165}$ & $1{\times}10^{-164}$ & 1095 \\
turbo    & 468 & 0 & $3{\times}10^{-141}$ & $3{\times}10^{-141}$ & 937 \\
large-v3 & 485 & 0 & $2{\times}10^{-146}$ & $4{\times}10^{-146}$ & 971 \\
\bottomrule
\end{tabular}
\caption{Exact paired McNemar tests \citep{mcnemar1947note} for stock versus
\texttt{eot\_row}@50 on all $630$ non-speech clips. $b$ is
stock-nonempty/patch-empty and $c$ the reverse. Holm correction is
over the six-scale family; OR uses the Haldane correction because every
$c=0$. All learned-patch LORs are $0.0\%$ (Wilson CI $[0.0,0.6]$).
The independently decoded analytic patch likewise has $c=0$ on every
scale and exact $p<10^{-130}$; its per-scale LORs are in
Table~\ref{tab:stats}. These clip-level tests are conditional on the
inspected battery and do not address method-selection or source-transfer
uncertainty.}
\label{tab:mcnemar}
\end{table}

\begin{table}[htbp]
\centering
\footnotesize
\setlength{\tabcolsep}{6pt}
\begin{tabular}{lrrr}
\toprule
Model & grid runs & max residual LOR & probe-WER range \\
\midrule
tiny     & 9 & 0.0 & $11.1$--$11.9$ \\
base     & 9 & 0.0 & $8.3$--$9.3$ \\
small    & 9 & 2.0 & $5.1$--$9.0$ \\
medium   & 9 & 0.0 & $3.9$--$63.1$ \\
turbo    & 9 & 0.0 & $2.4$--$11.4$ \\
large-v3 & 9 & 0.0 & $2.2$--$3.7$ \\
\bottomrule
\end{tabular}
\caption{Matched-capacity random-subnetwork verification on all six scales:
three learning rates by three mask seeds per scale. Max residual LOR is the
worst final normalized lexical-output rate across the nine runs. Reachability is
scale-wide (at most $2\%$), but WER safety is not: medium is acutely
learning-rate-sensitive. This supports no privileged parameter locus, not
interchangeable operating points.}
\label{tab:lrgrid}
\end{table}

\begin{table}[htbp]
\centering
\footnotesize
\begin{tabular}{rrrr}
\toprule
$p$ (non-speech) & $\mathrm{logit}(p)$ & abstain margin & $\|\Delta\|$ row \\
\midrule
0.20 & $-1.39$ & 12.67 & 0.359 \\
0.35 & $-0.62$ & 14.98 & 0.379 \\
0.50 & $+0.00$ & 14.46 & 0.373 \\
0.65 & $+0.62$ & 14.63 & 0.366 \\
0.80 & $+1.39$ & 16.10 & 0.353 \\
\bottomrule
\end{tabular}
\caption{Raw base-rate sweep behind Table~\ref{tab:baserate-replicated} (Whisper
small): converged step-0 abstain margin and unembedding row-delta norm at
each training non-speech fraction $p$. Fitted slope $0.98$ in
$\mathrm{logit}(p)$ ($r=0.85$; $95\%$ CI $[-0.12,2.09]$, $n{=}5$); the row
norm stays nearly constant; the row controls the gain while the base rate
changes the threshold.}
\label{tab:baseratepoints}
\end{table}

\begin{table}[htbp]
\centering
\footnotesize
\begin{tabular}{lrrrrrr}
\toprule
 & \multicolumn{3}{c}{Stock margin} & \multicolumn{3}{c}{Patched margin} \\
\cmidrule(lr){2-4}\cmidrule(lr){5-7}
Model & 0 & 10 & 30 & 0 & 10 & 30 \\
\midrule
tiny     & $-6.4$ & $-1.3$ & $-3.4$ & $+14.5$ & $+6.5$ & $+4.9$ \\
base     & $-6.2$ & $-1.0$ & $-2.6$ & $+15.2$ & $+6.1$ & $+5.9$ \\
small    & $-7.8$ & $-3.1$ & $-2.5$ & $+16.2$ & $+4.1$ & $+5.8$ \\
medium   & $-8.0$ & $-3.3$ & $-2.5$ & $+18.9$ & $+5.4$ & $+10.2$ \\
turbo    & $-4.6$ & $-3.4$ & $-4.9$ & $+12.0$ & $+2.2$ & $-0.8$ \\
large-v3 & $-4.9$ & $+0.3$ & $-8.0$ & $+9.8$  & $+6.3$ & $-4.8$ \\
\bottomrule
\end{tabular}
\caption{Position-wise abstain margin
$\mathrm{logit}(\mathrm{EOT}) - \max_t
\mathrm{logit}(\mathrm{content}_t)$ at emitted positions $0$, $10$, $30$, from
teacher-forcing each stock model on its own hallucinated transcript.
Stock margins are negative almost everywhere; the patch lifts them most at
early positions, and residual late negativity on turbo and large-v3 explains
why the position-0 intervention does not transfer directly to long-form
decoding.}
\label{tab:ratchet}
\end{table}

\begin{table}[htbp]
\centering
\footnotesize
\setlength{\tabcolsep}{3pt}
\begin{tabular}{lrrrr}
\toprule
Model & $\Delta$Sub & $\Delta$Del & $\Delta$Ins & $\Delta$WER \\
\midrule
tiny     & $-16.1$ & $+24.3$ & $-17.7$ & $-9.5$ \\
base     & $-7.8$  & $+12.4$ & $-13.7$ & $-9.1$ \\
small    & $-3.8$  & $+10.0$ & $-0.8$  & $+5.5$ \\
medium   & $-3.8$  & $+13.4$ & $-1.3$  & $+8.4$ \\
turbo    & $-2.9$  & $+12.4$ & $-0.7$  & $+8.8$ \\
large-v3 & $-3.0$  & $+15.1$ & $-0.6$  & $+11.6$ \\
\bottomrule
\end{tabular}
\caption{Fixed-subset decomposition of the degraded-speech WER change
under the trained EOT row, in percentage points. Tiny and base remove
enough substitutions and insertions to offset added deletions; the four
stronger scales over-delete.}
\label{tab:ledger}
\end{table}

\begin{table}[htbp]
\centering
\scriptsize
\setlength{\tabcolsep}{3pt}
\begin{tabular}{lrrr}
\toprule
Variant & gap words/min & deletions & WER \\
\midrule
stock & 62.0 & 2.5 & 7.4 \\
\texttt{eot\_row}@50 & 27.6 & 14.5 & 21.1 \\
VAD mute $+$ stock & 81.2 & 2.2 & 16.0 \\
VAD segmentation $+$ stock & \textbf{4.7} & \textbf{0.3} & \textbf{4.5} \\
boundary sweep & $26.3$--$72.9$ & $1.1$--$15.6$ & $22.0$--$54.6$ \\
\bottomrule
\end{tabular}
\caption{Whisper turbo on 40 clips (38.7 gap-minutes). The earlier
$81.2$ row muted gaps but decoded fixed windows; it was mislabeled as
segmentation. Correct VAD segmentation cuts and independently decodes
speech regions, dominating the null patch on both frontier axes. Rates are
percent except gap words/min.}
\label{tab:longform-corrected}
\end{table}

\begin{table}[htbp]
\centering
\footnotesize
\begin{tabular}{l ccc ccc}
\toprule
 & \multicolumn{3}{c}{LayerNorm @ best-zero step}
 & \multicolumn{3}{c}{\texttt{eot\_row} @ 50} \\
\cmidrule(lr){2-4}\cmidrule(lr){5-7}
Model & WER & deletions & empty$/100$ & WER & deletions & empty$/100$ \\
\midrule
tiny     & $23.7\to50.9$ & $5.7\to41.8$ & 20 & $23.7\to63.1$ & $5.7\to55.6$ & 22 \\
base     & $16.5\to21.4$ & $4.2\to9.9$  & 7  & $16.5\to20.6$ & $4.2\to9.6$  & 2  \\
small    & $13.2\to23.4$ & $5.1\to15.3$ & 10 & $13.2\to23.1$ & $5.1\to16.6$ & 5  \\
medium   & $11.7\to12.1$ & $3.8\to4.2$  & 3  & $11.7\to14.9$ & $3.8\to8.0$  & 0  \\
turbo    & $13.8\to25.0$ & $4.0\to15.0$ & 30 & $13.8\to13.4$ & $4.0\to4.3$  & 27 \\
large-v3 & $12.3\to24.5$ & $3.4\to18.9$ & 62 & $12.3\to16.3$ & $3.4\to9.0$  & 25 \\
\bottomrule
\end{tabular}
\caption{Spontaneous-speech cost and checkpoint-selection risk.
Earnings-22 \citep{delrio2022earnings22} WER and deletion rates
(stock$\to$patched, \%) and empty outputs per 100 spontaneous MUSAN-speech
clips, for the LayerNorm regime at its best zero-hallucination checkpoint and
\texttt{eot\_row} at step 50. Clean-probe checkpoint selection is blind to
this: late checkpoints leave up to $62\%$ of spontaneous speech empty
(large-v3, LayerNorm at step 350); tiny pays $+27$--$+39$ pp WER under
any zeroing checkpoint; medium (LayerNorm) and turbo (\texttt{eot\_row}) are
nearly free.}
\label{tab:earnings}
\end{table}

\FloatBarrier
\FloatBarrier

\section{Cross-Lingual and Cross-Architecture Detail}

This appendix carries the held-out-language detail behind
Section~\ref{sec:class}, the language-identification repair
(Table~\ref{tab:langid}), and standard MT quality under EOS bias
(Table~\ref{tab:mtquality}). Unexecuted cross-architecture baseline
placeholders have been removed rather than presented as evidence.

\begin{table}[htbp]
\centering
\scriptsize
\setlength{\tabcolsep}{1.5pt}
\begin{tabular}{llrrr}
\toprule
Lang & split & stock & patched & WER \\
\midrule
en & pilot & 63 & 1.3 & 6.3 \\
ru & pilot & 100 & 0.8 & n.a. \\
de & pilot & 81 & 1.0 & 13.9 \\
ja & pilot & 100 & 2.0 & n.a. \\
hi & pilot & 86 & 0.2 & n.a. \\
zh & pilot & 100 & 0.7 & n.a. \\
ar & pilot & 100 & 0.5 & n.a. \\
\midrule
fr & held out & 18 & 1.5 & 19.7 \\
es & held out & 90 & 1.7 & 8.5 \\
it & held out & 80 & 1.7 & 10.8 \\
nl & held out & 36 & 1.8 & 24.5 \\
pt & held out & 90 & 3.0 & 10.6 \\
pl & held out & 92 & 2.5 & 24.1 \\
tr & held out & 81 & 3.0 & 20.3 \\
ko & held out & 92 & 13.3 & n.a. \\
\bottomrule
\end{tabular}
\caption{Whisper-small multilingual EOT row: hallucination \% on seven
calibration tags and eight held-out tags, plus real-speech WER where
word-based scoring was available. Seven held-out tags remain at
$1.5$--$3.0\%$; Korean is the visible $13.3\%$ exception. This is
multilingual calibration followed by language holdout, not English
zero-shot transfer.}
\label{tab:enpatch}
\end{table}

\begin{table}[htbp]
\centering
\footnotesize
\setlength{\tabcolsep}{4pt}
\begin{tabular}{llcc}
\toprule
Lang & Script & langid (stock$\to$fix) & WER (stock$\to$fix) \\
\midrule
en & Latin      & $0.38\to\textbf{1.00}$ & $0.048\to0.048$ \\
de & Latin      & $0.28\to\textbf{0.96}$ & $0.116\to0.116$ \\
fr & Latin      & $0.59\to\textbf{0.97}$ & $0.170\to0.170$ \\
es & Latin      & $0.19\to\textbf{0.91}$ & $0.081\to0.081$ \\
it & Latin      & $0.06\to\textbf{0.80}$ & $0.124\to0.124$ \\
nl & Latin      & $0.56\to\textbf{0.95}$ & $0.188\to0.188$ \\
pt & Latin      & $0.47\to\textbf{0.91}$ & $0.101\to0.101$ \\
pl & Latin      & $0.32\to\textbf{0.97}$ & $0.202\to0.202$ \\
tr & Latin      & $0.07\to\textbf{0.90}$ & $0.131\to0.131$ \\
ru & Cyrillic   & $0.32\to\textbf{0.89}$ & $0.141\to0.141$ \\
ar & Arabic     & $0.51\to\textbf{0.95}$ & $0.332\to0.332$ \\
hi & Devanagari & $0.22\to\textbf{0.96}$ & $0.654\to0.654$ \\
ja & Kana       & $0.06\to\textbf{0.97}$ & n.a.$^{\dagger}$ \\
ko & Hangul     & $0.06\to\textbf{0.92}$ & $0.225\to0.225$ \\
zh & Han        & $0.24\to\textbf{0.98}$ & n.a.$^{\dagger}$ \\
\bottomrule
\end{tabular}
\caption{One row per language token restores Whisper small's short-clip
($2$ s) language identification without touching transcription: langid
accuracy and word WER (stock$\to$fix) across fifteen FLEURS languages and
seven scripts. Mean langid rises $0.29\to0.94$; WER is bit-identical (the
edit is scoped to the detection step). $^{\dagger}$ja/zh word-WER is
$\approx 1$ by construction (no word boundaries); only its invariance under
the fix matters.}
\label{tab:langid}
\end{table}

\begin{table}[htbp]
\centering
\footnotesize
\setlength{\tabcolsep}{1.5pt}
\begin{tabular}{llcc}
\toprule
Model & Intervention & Halluc.\ (\%) & WER \\
\midrule
wav2vec2 (CTC)      & anchor FT     & $14.4\to\mathbf{0.8}$  & $2.30\to2.30$ \\
HuBERT (CTC)        & anchor FT     & $61.4\to\mathbf{0.5}$  & $2.24\to4.66$ \\
WavLM (CTC)         & bias $\beta{=}2$ & $36.2\to\mathbf{14.0}$ & $9.15\to10.77$ \\
wav2vec2-base (CTC) & bias $\beta{=}2$ & $30.8\to\mathbf{16.8}$ & $4.15\to5.72$ \\
Conformer (CTC)     & bias $\beta{=}2$ & $11.1\to\mathbf{4.3}$  & $2.75\to3.82$ \\
Emformer (RNN-T)    & bias $\beta{=}4$ & $24.8\to\mathbf{7.3}$  & $7.13\to8.14$ \\
Conformer (RNN-T)   & none needed   & $\mathbf{1.9}$         & $1.65$ \\
\bottomrule
\end{tabular}
\caption{The blank is the CTC/RNN-T null token. Hallucination and WER are
stock$\to$patched (\%); ``bias'' is a single blank-logit scalar. Strong CTC
systems reach near-zero via anchor fine-tuning; weaker systems reduce
monotonically under one scalar.}
\label{tab:blank}
\end{table}

\begin{table}[htbp]
\centering
\footnotesize
\setlength{\tabcolsep}{5pt}
\begin{tabular}{lrrrr}
\toprule
Released AudioSAE-small condition & HR & $95\%$ CI & WER & $\Delta$WER \\
\midrule
stock (SAE disabled)       & 96.2 & $[94.4,97.4]$ & 3.73   & $+0.00$ \\
reconstruction ($\alpha=0$)& 96.2 & $[94.4,97.4]$ & 3.41   & $-0.32$ \\
steering $\alpha=1$        & 94.6 & $[92.6,96.1]$ & 3.74   & $+0.01$ \\
steering $\alpha=2$        & 95.1 & $[93.1,96.5]$ & 61.33  & $+57.60$ \\
steering $\alpha=4$        & 48.9 & $[45.0,52.8]$ & 188.57 & $+184.85$ \\
steering $\alpha=8$        & 0.0  & $[0.0,0.6]$   & 100.00 & $+96.27$ \\
\bottomrule
\end{tabular}
\caption{Common-axis evaluation of the released
AudioSAE-Whisper-small final-layer encoder SAE
\citep{audiosae2026}. We select the top 128 features by a logistic
regression fit only on training pools and apply the released steering
recipe. At the paper-strength setting $\alpha=1$, HR barely moves; increasing
strength destroys transcription before suppressing fabrication. Zero HR is
reached only when WER is $100\%$. One SAE layer replicated across the twelve
encoder layers contributes $113.3$M parameters, versus $768$ for the EOT
row.}
\label{tab:audiosae}
\end{table}

\begin{table}[htbp]
\centering
\footnotesize
\setlength{\tabcolsep}{5pt}
\begin{tabular}{lllrrr}
\toprule
Model & Method & Screening & Params & HR & WER \\
\midrule
Canary & stock & -- & -- & 77.5 & 1.66 \\
Canary & Calm top-8 head ablation & 192 passes & 0 & 72.9 & 1.70 \\
Canary & EOS scalar, $\beta=8$ & none & 1 & 0.0 & 6.94 \\
\midrule
OWSM & stock & -- & -- & 100.0 & 16.11 \\
OWSM & Calm top-8 head ablation & 288 passes & 0 & 100.0 & 16.08 \\
OWSM & EOS scalar, $\beta=9$ & none & 1 & 4.7 & 12.79 \\
\bottomrule
\end{tabular}
\caption{Cross-architecture Calm-Whisper route versus a one-scalar EOS
edit. Screening all candidate decoder heads finds a maximum step-0
$\Delta P(\mathrm{EOS})$ of only $0.0006$ on Canary and $0.0001$ on OWSM;
ablating the selected heads consequently has little or no effect. Canary
uses $630$ non-speech clips and $500$ test-clean utterances. The slower OWSM
ablation uses $300/150$ with \texttt{maxlenratio=0.4}, so its absolute WER
is not comparable with other OWSM tables, but its stock/ablation comparison
is paired. Scalar rows report their independently measured operating
points.}
\label{tab:calmxarch}
\end{table}

\begin{table}[htbp]
\centering
\footnotesize
\setlength{\tabcolsep}{6pt}
\begin{tabular}{rrr}
\toprule
OWSM EOS bias $\beta$ & HR & WER \\
\midrule
0.0  & 90.7 & 8.19 \\
2.0  & 84.0 & 8.19 \\
4.0  & 72.7 & 8.19 \\
6.0  & 58.0 & 8.92 \\
8.0  & 16.7 & 9.43 \\
8.5  & 9.3  & 11.04 \\
9.0  & 4.7  & 12.79 \\
9.5  & 4.0  & 15.13 \\
10.0 & 1.3  & 20.91 \\
11.0 & 0.0  & 48.76 \\
12.0 & 0.0  & 89.69 \\
\bottomrule
\end{tabular}
\caption{Fine OWSM dose response on the fixed $150$ non-speech plus $60$
speech subset. The bias is a monotone hallucination control, but there is
no free operating point: the empirical balance point is $\beta=9$, whereas
the $1.3\%$ HR headline at $\beta=10$ more than doubles WER.}
\label{tab:owsmgrid}
\end{table}

\begin{table}[htbp]
\centering
\footnotesize
\setlength{\tabcolsep}{6pt}
\begin{tabular}{llccccc}
\toprule
Model & $\beta$ & BLEU & chrF & COMET & nonempty & length \\
\midrule
NLLB en$\to$ru & $0$  & 29.0 & 54.4 & 83.4 & 100 & 91 \\
NLLB en$\to$ru & $6$  & 25.2 & 50.8 & 80.9 & 100 & 82 \\
NLLB en$\to$ru & $8$  & 13.9 & 37.4 & 70.3 & 100 & 54 \\
\midrule
Marian en$\to$de & $0$  & 37.7 & 63.7 & 79.9 & 100 & 97 \\
Marian en$\to$de & $6$  & 34.5 & 59.0 & 75.6 & 100 & 87 \\
Marian en$\to$de & $10$ & 0.0  & 5.9  & 32.2 & 87  & 8 \\
\bottomrule
\end{tabular}
\caption{Real-source WMT19 translation quality under EOS bias. BLEU
\citep{papineni2002bleu}, chrF \citep{popovic2015chrf}, and COMET
\citep{rei2020comet} (\texttt{wmt22-comet-da}, multiplied by $100$) decline
at the working point ($\beta=6$) and decline sharply under stronger
early-termination bias.
Nonempty and length are percentages.}
\label{tab:mtquality}
\end{table}

\FloatBarrier
\FloatBarrier

\section{Cross-Architecture Logit-Lens Detail}
\label{sec:logitlens-detail}

\begin{figure}[!ht]
\centering
\includegraphics[width=0.6\columnwidth,height=0.82\textheight,keepaspectratio]{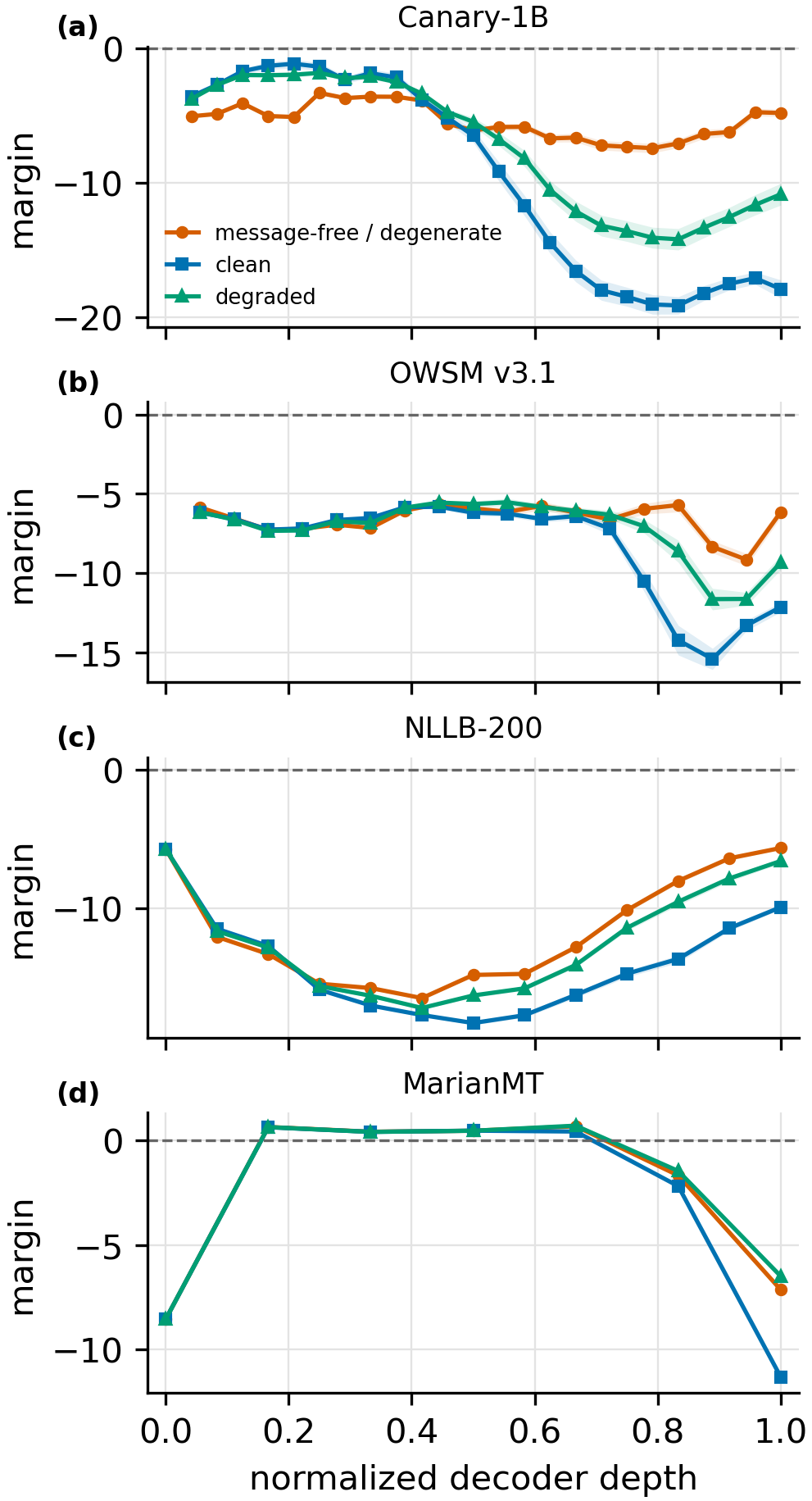}
\caption{Cross-architecture logit-lens depth traces at decode step~0. Curves
show the mean EOS abstain margin for message-free audio or degenerate source
text, clean input, and degraded input; bands are normal-approximation 95\%
confidence intervals over $n=120$ clips per condition for Canary/OWSM and
$n=200$ sources per condition for NLLB/Marian. The dashed line is the EOS
argmax boundary. Message-free input finishes above clean input in every
model, but below zero. Marian uniquely promotes EOS in the middle stack and
demotes it in the final block, delimiting the stronger claim that abstention
is never internally promoted.}
\label{fig:logitlens-crossarch}
\end{figure}

Figure~\ref{fig:logitlens-crossarch} expands the final-margin summary in
Figure~\ref{fig:logitlens-summary}b into complete decoder-depth traces for
the two non-Whisper speech AEDs and two translation models. Canary, OWSM,
and NLLB keep the mean null-token margin below zero throughout. Marian is
qualitatively different: EOS is the mean argmax through the middle stack
for all three source conditions, but the last decoder block reverses that
preference, most strongly on clean sources. The common final failure is
therefore compatible with different internal trajectories. Three models do
not promote abstention, while Marian demotes it late. We consequently focus
on the final readout decision rather than claim one universal layer-wise
mechanism.

\end{document}